\documentclass{article}

    \PassOptionsToPackage{numbers, compress}{natbib}
    \usepackage[final]{neurips_2025}

\usepackage[utf8]{inputenc} % allow utf-8 input
\usepackage[T1]{fontenc}    % use 8-bit T1 fonts
\usepackage{hyperref}       % hyperlinks
\usepackage{url}            % simple URL typesetting
\usepackage{booktabs}       % professional-quality tables
\usepackage{amsfonts}       % blackboard math symbols
\usepackage{nicefrac}       % compact symbols for 1/2, etc.
\usepackage{microtype}      % microtypography
\usepackage{xcolor}         % colors

\usepackage{amsmath}
\usepackage{amssymb}
\usepackage{graphicx}
\usepackage{xspace}
\usepackage{cleveref}
\usepackage{multirow}
\usepackage{threeparttable}
\usepackage{adjustbox}
\usepackage[table]{xcolor}
\usepackage{colortbl}
\usepackage{wrapfig}
\usepackage{caption}

\makeatletter
\DeclareRobustCommand\onedot{\futurelet\@let@token\@onedot}
\def\@onedot{\ifx\@let@token.\else.\null\fi\xspace}

\def\eg{\emph{e.g}\onedot} 
\def\ie{\emph{i.e}\onedot} 
 
\def\etc{\emph{etc}\onedot}

\def\etal{\emph{et al}\onedot}
\makeatother

\definecolor{first}{HTML}{b7c29e}
\definecolor{second}{HTML}{d3dac5}
\definecolor{third}{HTML}{f0f1ec}
\definecolor{G28}{RGB}{71, 98, 5}
\definecolor{G28l}{RGB}{181, 192, 155}

\newcommand{\mytoprule}{\specialrule{0.8pt}{0em}{0em}}
\newcommand{\mymidrule}{\specialrule{0.6pt}{0em}{0em}}
\newcommand{\mybottomrule}{\specialrule{0.8pt}{0em}{0em}}
\newcommand{\mydmidrule}{\specialrule{0.6pt}{0em}{1pt}\specialrule{0.6pt}{0em}{0em}}
\newcommand{\mycmidrule}[1]{\noalign{\global\arrayrulewidth=0.6pt}\cline{#1}\noalign{\global\arrayrulewidth=0.4pt}}

\title{Re-coding for Uncertainties: Edge-awareness Semantic Concordance for Resilient Event-RGB Segmentation}

\author{%
  Nan Bao$^{1}$,\quad Yifan Zhao$^{1}$,\quad Lin Zhu$^{2}$,\quad Jia Li$^{1}$\thanks{Correspondence should be addressed to Jia Li. Website: \texttt{https://cvteam.buaa.edu.cn/}} \\
  $^{1}$State Key Laboratory of Virtual Reality Technology and Systems, SCSE \& QRI, Beihang University \\
  $^{2}$School of Computer Science and Technology, Beijing Institute of Technology \\
  \texttt{\{nbao, zhaoyf, jiali\}@buaa.edu.cn,\quad linzhu@bit.edu.cn}
}

\begin{document}

\maketitle

\begin{abstract}

Semantic segmentation has achieved great success in ideal conditions.
However, when facing extreme conditions (\eg, insufficient light, fierce camera motion), most existing methods suffer from significant information loss of RGB, severely damaging segmentation results.
Several researches exploit the high-speed and high-dynamic event modality as a complement, but event and RGB are naturally heterogeneous, which leads to feature-level mismatch and inferior optimization of existing multi-modality methods.
Different from these researches, we delve into the edge secret of both modalities for resilient fusion and propose a novel Edge-awareness Semantic Concordance framework to unify the multi-modality heterogeneous features with latent edge cues.
In this framework, we first propose Edge-awareness Latent Re-coding, which obtains uncertainty indicators while realigning event-RGB features into unified semantic space guided by re-coded distribution, and transfers event-RGB distributions into re-coded features by utilizing a pre-established edge dictionary as clues.
We then propose Re-coded Consolidation and Uncertainty Optimization, which utilize re-coded edge features and uncertainty indicators to solve the heterogeneous event-RGB fusion issues under extreme conditions.
We establish two synthetic and one real-world event-RGB semantic segmentation datasets for extreme scenario comparisons.
Experimental results show that our method outperforms the state-of-the-art by a 2.55\% mIoU on our proposed DERS-XS, and possesses superior resilience under spatial occlusion.
Our code and datasets are publicly available at \texttt{https://github.com/iCVTEAM/ESC}.

\end{abstract}
\section{Introduction}

Being widely used in autonomous driving, medical imaging, geospatial analysis, and industrial inspection, semantic segmentation aims to resolve the semantics of visual objects, assigning category labels to each pixel in the image \cite{chen2023generative}.
When facing extreme conditions due to diversity and complexity in the wild, conventional single-RGB semantic segmentation faces challenges of corrupted results, suffering from significant information loss. This has led to the exploration of leveraging information from multiple modalities for semantic segmentation \cite{xie2024eisnet, zhang2023cmx, zhang2023delivering}.

\begin{wrapfigure}{r}{0.67\textwidth}
    \centering
    \vspace{-0.5em}
    \includegraphics[width=1\linewidth]{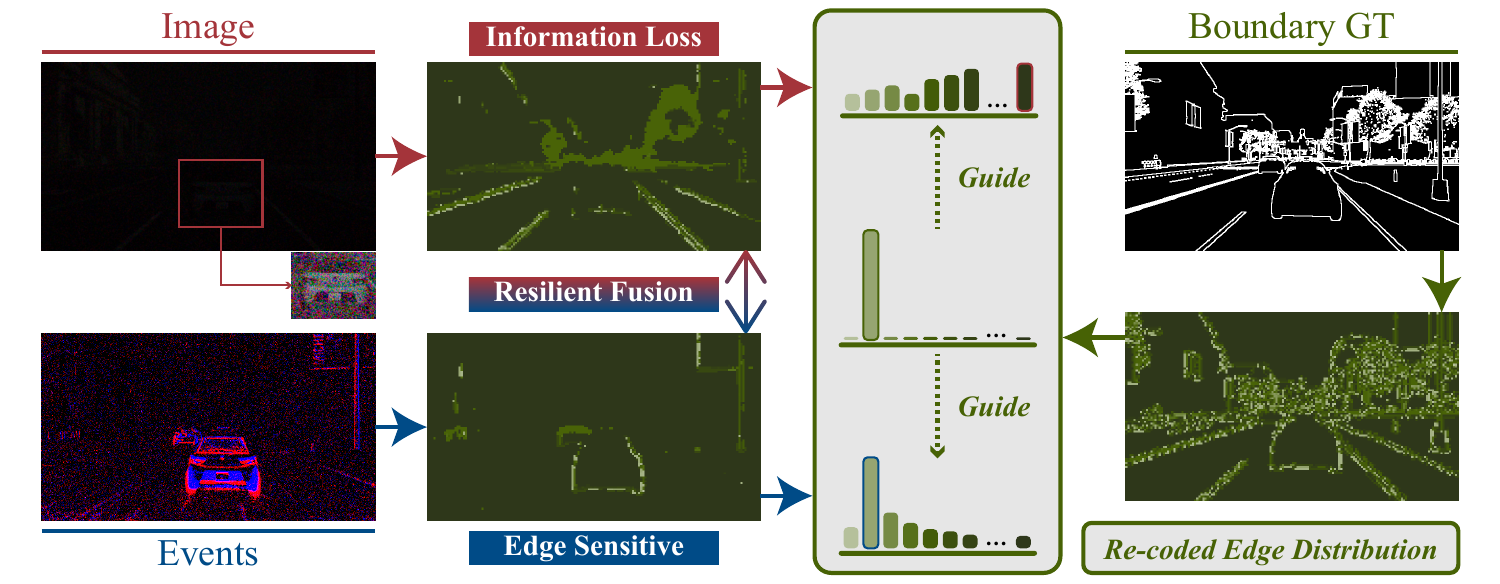}
    \captionsetup{font=small}
    \caption{
    \textbf{Edge-awareness Semantic Concordance for event-RGB fusion.} RGB suffers from severe information loss under extreme conditions, while events are sensitive to edges in motion, complementing the lost information. Heterogeneous properties of event and RGB lead to feature-level mismatch and inferior optimization of existing methods. Our ESC framework utilizes semantic edge as an intermediate commonality for a more resilient fusion.
    }
    \vspace{-0.5em}
    \label{fig:Head}
\end{wrapfigure}

We investigate the problem of leveraging event and RGB for semantic segmentation under extreme conditions, focusing on inferior optimization issues in modality imbalance and failure situations.
Dynamic vision sensor \cite{lichtsteiner2008128, brandli2014240, taverni2018front}, commonly known as event camera, responds to brightness changes and generates events for each pixel asynchronously and independently. This unique mechanism gives it many advantages beyond conventional cameras, such as high dynamic range, high temporal resolution, low latency, and low power consumption \cite{gallego2020event}. Therefore, event data are widely used in tasks that are hard to solve with conventional images alone, such as HDR image reconstruction \cite{rebecq2019high, stoffregen2020reducing, zhang2020learning}, motion deblurring \cite{pan2019bringing, jiang2020learning, shang2021bringing}, and low-light enhancement \cite{jiang2023event, liang2023coherent, zhou2021delieve, zhou2023deblurring}. As shown in \cref{fig:Head}, image suffers from severe information loss under extreme conditions due to a low signal-to-noise ratio, while events clearly show the motion edge of vehicles. It becomes feasible to complement the lost information of RGB modality by utilizing event modality.

However, existing event-RGB semantic segmentation methods \cite{zhang2023cmx, zhang2023delivering, xie2024eisnet} do not consider the heterogeneous properties of event and RGB modality. Therefore, although the naive fusion strategy has achieved some improvements, it is difficult to handle feature-level mismatch and inferior optimization issues, especially in modality imbalance and failure situations. To overcome the above problems of heterogeneous event and RGB, we find semantic edge as an intermediate commonality for both. 

Existing studies \cite{li2020improving, zhen2020joint, xiao2023baseg} have proven that edge-awareness is beneficial for RGB segmentation. Intuitively, events highlight edges in motion, and RGB gradients reveal edge cues. Through statistics in \cref{sec:3.1}, we indeed find a strong correlation between events and semantic edge. \textbf{Semantic edge serves as a bridge, guiding the heterogeneous event and RGB to embed into a unified semantic space.} By utilizing information of semantic edge as bridge, we successfully realign the heterogeneous event and RGB into the unified semantic space to jointly optimize their edge information, while consolidating the image contextual information with semantic edge information as crucial clue.

In this paper, we propose Edge-awareness Semantic Concordance (ESC), a novel multi-modality learning framework for event-RGB semantic segmentation. 
ESC utilizes a shared discrete embedding space, creating an edge dictionary containing basic semantic elements from semantic edge. 
We introduce Edge-awareness Latent Re-coding for discrete latent space modeling and transferring bi-directionally, namely re-coding. 
The re-coded edge features are utilized for information consolidation, and the re-coded edge distribution enables unified realignment through cross-entropy supervision.
Uncertainty indicators are derived from modality distributions for joint optimization.
Re-coded Consolidation and Uncertainty Optimization are designed to achieve the above processes for resilient fusion.
Prior work \cite{xie2024eisnet} assesses event-RGB segmentation using RGB-pseudo-labeled datasets (\eg, DDD17 \cite{binas2017ddd17}, DSEC-Semantic\cite{gehrig2021dsec, sun2022ess}), leading to potentially unreliable results.
To address this, we introduce synthetic DERS-XS and real-world DERS-XR, featuring low-light RGB, noisy events, and true-labels for extreme scenario comparisons.
We further adapt DSEC-Semantic into an extreme variant, DSEC-Xtrm, to mitigate direct dependence of pseudo-labels on original RGB.
Experiments on above datasets show that our method achieves better performance and is more resilient under extreme conditions compared to existing multi-modality methods.
To the best of our knowledge, this is the first work to assess model resilience via spatial occlusion evaluation without any fine-tuning.

The contributions of our work are summarized as follows:

1) We propose Edge-awareness Semantic Concordance (ESC), a novel multi-modality framework that exploits supervision over re-coded distribution to realign heterogeneous event and RGB into unified semantic space, jointly optimizing them based on uncertainties derived from modality distributions. 

2) We propose three modules, namely Edge-awareness Latent Re-coding (ELR), Re-coded Consolidation (RC), and Uncertainty Optimization (UO). ELR re-codes features and distributions bi-directionally, while RC and UO utilize the re-coded features and uncertainties for a resilient fusion.

3) We establish two synthetic and one real-world event-RGB semantic segmentation datasets for extreme scenario comparisons. Experimental results show that our method outperforms the state-of-the-art methods and possesses superior resilience under extreme conditions including occlusion.
\section{Related work}

\subsection{Event-based semantic segmentation}

Event data has recently been applied to semantic segmentation tasks. Ev-SegNet \cite{alonso2019ev} introduces semantic segmentation for event data by proposing a 6-channel image-like representation and applying CNN architectures on the DDD17 dataset \cite{binas2017ddd17}. EvDistill \cite{wang2021evdistill} trains a student network on unlabeled event data via knowledge distillation from a large image-based teacher network. 
CMDA \cite{DBLP:conf/iccv/XiaZZWST23} proposes an unsupervised domain adaption segmentation framework to transfer daytime RGB knowledge to nighttime event domain. ESS \cite{sun2022ess} leverages labeled images for training on unlabeled event data through unsupervised domain adaptation. EvSegFormer \cite{jia2023event} introduces a posterior attention module to utilize prior knowledge from event data, and HPL-ESS \cite{jing2024hpl} proposes a hybrid pseudo-labeling framework to mitigate noisy labels in unsupervised event-based segmentation. 
ESEG \cite{zhao2025eseg} is a uni-modality event-based segmentation framework that exploits edge semantics to provide explicit guidance toward the regions of interest.
ISSAFE \cite{zhang2021issafe} leverages events to assist segmentation under accident scenes, and HALSIE \cite{biswas2024halsie} features a hybrid dual-encoder scheme with SNN and ANN for efficient segmentation. 
Recent works demonstrate the feasibility of leveraging event data, while most works do not fully explore its unique characteristics, limiting its advantages over conventional RGB.

\subsection{Event-assisted vision tasks}

Event data can be utilized for assisting conventional vision tasks due to its high-speed and high-dynamic capacity. 
Pan \etal \cite{pan2019bringing} propose an event-based double integral model to restore sharp video from a single blurry frame with events.
Jiang \etal \cite{jiang2020learning} recover sharp videos with events based on a recurrent neural network. Shang \etal \cite{shang2021bringing} utilize events for non-consecutively frames deblurring. 
Liang \etal \cite{liang2023coherent} and Liu \etal \cite{liu2023low} leverage event data to guide low-light video enhancement. 
Jiang \etal \cite{jiang2023event} propose a joint framework to reconstruct clear images from underexposed frames and event streams. 
Shi \etal \cite{shi2023even} utilize paired images and event streams to estimate monocular depth under night conditions. 
Li \etal \cite{li2024event} propose an event-based low-light video object segmentation framework. 
Qi \etal \cite{qi2023e2nerf} introduce events into neural radiance fields for novel view sharp image rendering.
Geng \etal \cite{geng2024event} introduce events into visible and infrared fusion task.
There are also several cross-modality contrastive pretraining approaches, such as Yang \etal \cite{yang2023event}, Yao \etal \cite{yao2024sam}, and Wu \etal \cite{wu2025cm3ae}, aiming to acquire informative and effective pretrained backbones for both event and RGB.
Prevailing research proves the effectiveness of event-assisted tasks, while the inferior optimization issue of heterogeneous event and RGB under extreme conditions remains unexplored.
            
\subsection{Inter-modality Fusion}

Inter-modality fusion is the core issue of multi-modality tasks. How to obtain better-fused features has become an enduring research topic. 
Zhang \etal \cite{zhang2023cmx, zhang2023delivering} aim to achieve a general cross-modality segmentation model for arbitrary modalities, including event modality. Xie \etal \cite{xie2024eisnet} propose a modality recalibration and fusion module to recalibrate and then aggregate events and image features at multiple stages. Other works only focus on fusion techniques without specifying a specific task. Wang \etal \cite{wang2022multimodal} detect tokens with less information dynamically and substitute them with aggregated features projected from another modality. Jia \etal \cite{DBLP:conf/icml/JiaG0W00024} introduces noise embeddings into proposed inter-modality attention module to improve interaction between features of multi-modality pixel-wise. 
Zhao \etal \cite{zhao2024edge} utilized extra edge cues for event-RGB stereo.
Several approaches have also emerged, including Kim \etal \cite{kim2024missing} and Lang \etal \cite{lang2025retrieval}, to address the problem of incomplete modality inputs.
In vision-language models, discrete representation learning with shared embedding space is becoming popular. Liu \etal \cite{liu2022cross} propose a representation learning paradigm that contains a discretized embedding space shared across two different modalities such as video and audio. Xia \etal \cite{xia2024achieving} propose a framework that obtains mutual semantic information from different modalities by modality feature reconstruction. Zheng \etal \cite{zheng2024unicode} propose an iterative learning paradigm for tuning large language models into multi-modality LLM. Zhou \etal \cite{DBLP:conf/iclr/Zhou0LYDSY24} draw on the concept of shared latent space and first introduce it into domain adaptation vision task of nighttime optical flow estimation.

Inspired by the above works, we propose an Edge-awareness Semantic Concordance framework to model a shared discrete latent edge space and optimize events and image features into the unified semantic space based on the re-coded edge distribution. By edge-awareness latent re-coding, we obtain re-coded edge features and uncertainties, which are utilized for inter-modality resilient optimization.
\section{Method}

\label{sec:3}

Naive fusion strategy fails to integrate heterogeneous event and RGB under extreme conditions.
We propose an Edge-awareness Semantic Concordance framework to address this. 
To prove the rationality, we first analyze event edge characteristics in \cref{sec:3.1}. 
We then establish an edge dictionary as a preliminary in \cref{sec:estabEdgedic}. 
Based on this dictionary, Edge-awareness Latent Re-coding (\cref{sec:ELR}) transforms edge distribution and features bi-directionally, namely re-coding. 
Re-coded edge distribution is utilized for feature-level unified realignment through supervision. 
Re-coded Consolidation and Uncertainty Optimization (\cref{sec:REC} and \cref{sec:UEJO}) utilizes re-coded edge features and uncertainties derived from modality distributions for a resilient fusion. 
Since labels from DSEC-Semantic are unreliable, we construct three new datasets for reliable evaluation in \cref{sec:datasets}. 

\subsection{Edge characteristic of events}
\label{sec:3.1}

\begin{wrapfigure}{r}{0.65\textwidth}
    \centering
    \vspace{-1.25em}
    \includegraphics[width=1.0\linewidth]{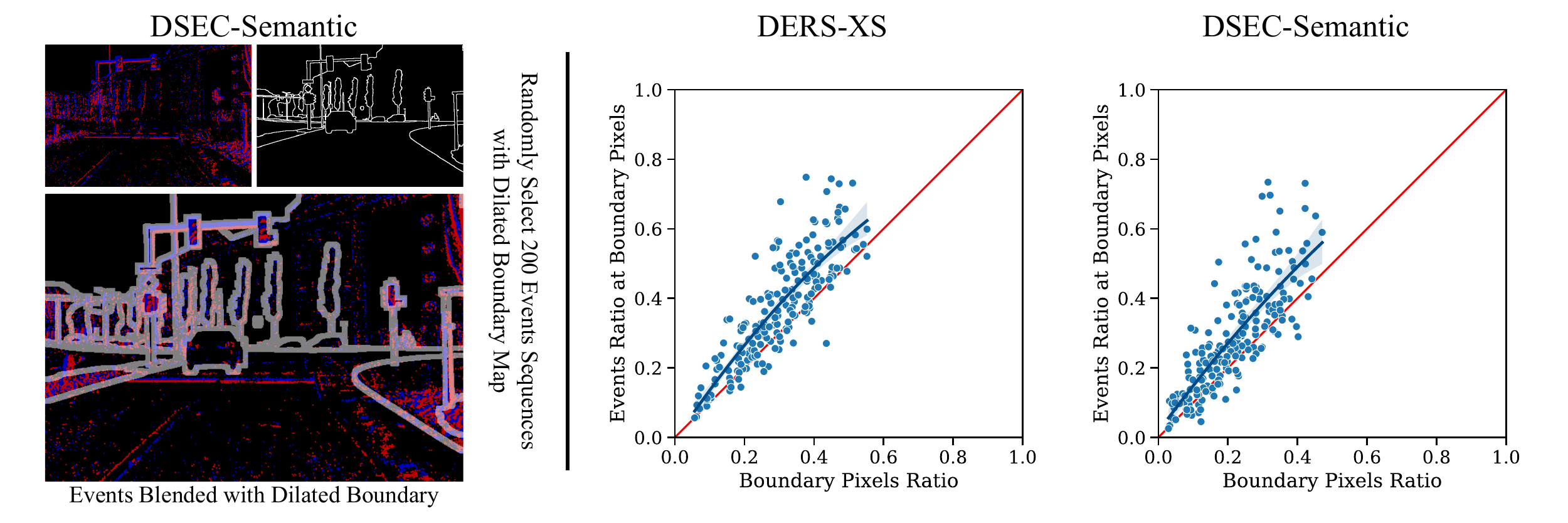}
    \captionsetup{font=small}
    \caption{\textbf{Correlation between events and semantic edge.} We randomly select 200 event sequences with dilated boundary map from true-labeled DERS-XS and real-world DSEC-Semantic, counting the ratio of edge pixels to all pixels and the ratio of events at edge pixels to all events, respectively. For both datasets, as the area of edge expands, the events ratio is always greater than the boundary ratio. This exhibits a strong correlation between events and semantic edge under different conditions.}
    \label{fig:EdgeRelation}
    \vspace{-0.75em}
\end{wrapfigure}

Event camera is a bio-inspired sensor that triggers event signals asynchronously when light intensity changes at each pixel. Specifically, as \cref{eq:eq3-1} shows, an event $\mathbf{e} = \langle \mathbf{x}, t, p_{\mathbf{x}, t} \rangle$ is triggered when pixel $\mathbf{x} = \langle x, y \rangle$ perceives a change in light intensity $I$ that reaches threshold $\Theta$ in the logarithmic domain at time $t$, where $p_{\mathbf{x}, t}$ means polarity of light intensity change in logarithm domain. Triggered events from $t_{start}$ to $t_{end}$ form an event stream $\{\mathbf{e}_i = \langle \mathbf{x}_i, t_i, p_i \rangle\}_{t_{start} < t_i \leq t_{end}}$.
\begin{equation}
    \begin{split}
        p_{\mathbf{x}, t} = \left \{ \begin{aligned} +1, \quad \log (I_{\mathbf{x}, t}) - \log (I_{\mathbf{x}, t - \Delta t}) &> \Theta, \\ -1, \quad \log (I_{\mathbf{x}, t}) - \log (I_{\mathbf{x}, t - \Delta t}) &< -\Theta.  \end{aligned} \right.
    \end{split}
    \label{eq:eq3-1}
\end{equation}

We demonstrate the correlation between events and semantic edge (\ie, segmentation boundary) through statistics in \cref{fig:EdgeRelation}. 
We randomly select 200 event sequences with dilated boundary map from true-labeled DERS-XS and real-world DSEC-Semantic, counting the ratio of edge pixels to all pixels of the whole plane and the ratio of events falling on edge pixels to all events of the whole sequence. Statistical results show that as the area of edge pixels expands, the events ratio is always greater than the boundary ratio for both datasets. This demonstrates that events tend to cluster at areas of semantic edge under different conditions, exhibiting a strong correlation between events and semantic edge, which supports our utilization of semantic edge as a bridge for heterogeneous event and RGB.
Details of statistical process of event-edge correlation with more analyses can be seen in \cref{sec:appendixC}.

\subsection{Edge dictionary as intermediate semantic clues across modalities}

\label{sec:estabEdgedic}

% NIPS CR Ver.
\begin{figure}[h!]
    \centering
    \vspace{-1.25em}
    \includegraphics[width=0.8\linewidth]{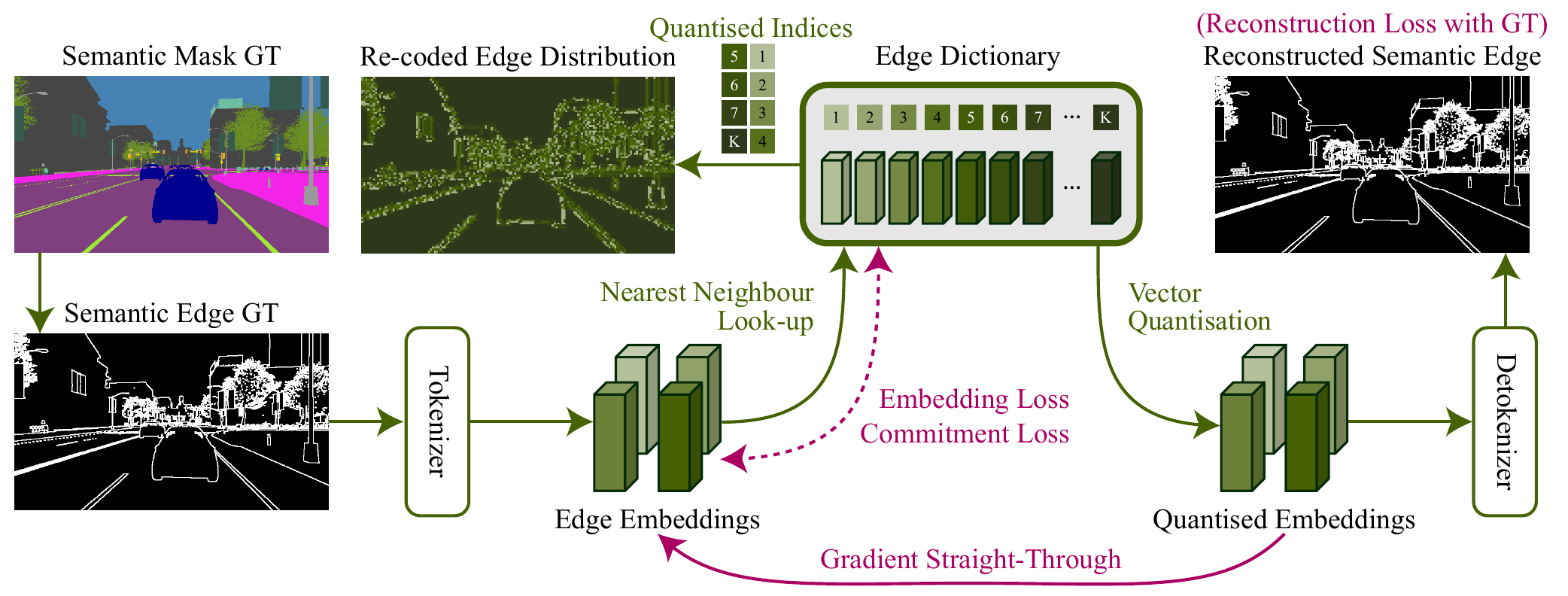}
    \captionsetup{font=small}
    \caption{\textbf{Establishment of edge dictionary.} We establish our edge dictionary based on a VQ-VAE architecture. Semantic edge is retrieved from the semantic mask ground truth and leveraged for learning its discrete latent representations as an edge dictionary, which serves as intermediate clues across heterogeneous event and RGB.}
    \label{fig:EdgeDic}
\end{figure}

To utilize semantic edge as intermediate clues, we establish an edge dictionary, which is a discrete latent embedding space derived from semantic edge, containing basic semantic elements of edge and shared by heterogeneous event and RGB. The establishment of our edge dictionary is shown in \cref{fig:EdgeDic} and based on a VQ-VAE \cite{van2017neural} architecture, which is originally used for representation learning and used by \cite{liu2022cross, ning2023all, xia2024achieving, zheng2024unicode} to model shared discrete latent space from inputs.

We first retrieve semantic edge from semantic mask ground-truth by a mean filter followed by an indicator function. Given a semantic mask $\mathbf{M} \in {\{1, \cdots, c\}}^{H \times W}$, boundary map $\mathbf{B} \in {\{0, 1\}}^{H \times W}$ can be obtained by $\mathbf{B} = {\mathbb{I}}_{\mathbf{M} \neq \text{Mean-Filter}(\mathbf{M})}(\mathbf{M})$, where $c$ is number of categories in semantic mask, and $\mathbb{I}$ is indicator function. 

We define edge dictionary as $\Delta = \{\langle k, v(k) \rangle|k \in \{1, \cdots, K\}\}$, where $K$ is the number of items (\ie quantised vectors) that edge dictionary contains, and $v(\cdot)$ is the query function as $v(k) \in \mathbb{R}^n$ selects the $k$-th item of edge dictionary with $n$-dimension. 
As \cref{fig:EdgeDic} shows, the tokenizer $f_T$ takes boundary map $\mathbf{B}$ as input, producing edge embeddings $\mathbf{\Gamma} = f_T(\mathbf{B}) \in \mathbb{R}^{H' \times W' \times n}$, which have downsampled spatial size $H' \times W'$ and $n$ channels. 
Items in edge dictionary are selected by nearest neighbour look-up method as $\mathbf{\Gamma}' = v(\hat{\mathbf{K}}) = v(\mathop{\arg \min}_{\mathbf{K} \in \{1, \cdots, K\}^{H'\times W'}}\|\mathbf\Gamma-v(\mathbf{K})\|_2^2)$, where $\hat{\mathbf{K}}$ contains the queried indices and $\mathbf{\Gamma}'$ is the quantised edge embeddings.
Reconstructed boundary map $\mathbf{B}'$ is obtained by $\mathbf{B}'=f_{T'}(\mathbf{\Gamma}')$, where $f_{T'}$ is the detokenizer.

To ensure the edge dictionary contains all basic information of semantic edge, we need to ensure the boundary map is reconstructed flawlessly while items in edge dictionary are close enough to edge embeddings. Thus, we adopt the training objective with reconstruction loss, embedding loss, and commitment loss as 
$L_{dict} = \|\mathbf{B}-\mathbf{B}'\|_2^2~+ \|v(\hat{\mathbf{K}}) - \mbox{sg}(\mathbf{\Gamma})\|_2^2+\alpha\|\mbox{sg}(v(\hat{\mathbf{K}})) - \mathbf{\Gamma}\|_2^2,$
where $\mbox{sg}(\cdot)$ means stop gradient, and $\alpha$ is a constant of commitment loss weight. To make the reconstruction loss propagate back to tokenizer, a gradient straight-through technique is adopted, which directly assigns the gradient from $\mathbf{\Gamma}'$ to $\mathbf{\Gamma}$.
Details of edge dictionary training process can be seen in \cref{sec:appendixD}.

\begin{figure*}[t]
    \centering
    \includegraphics[width=1.0\linewidth]{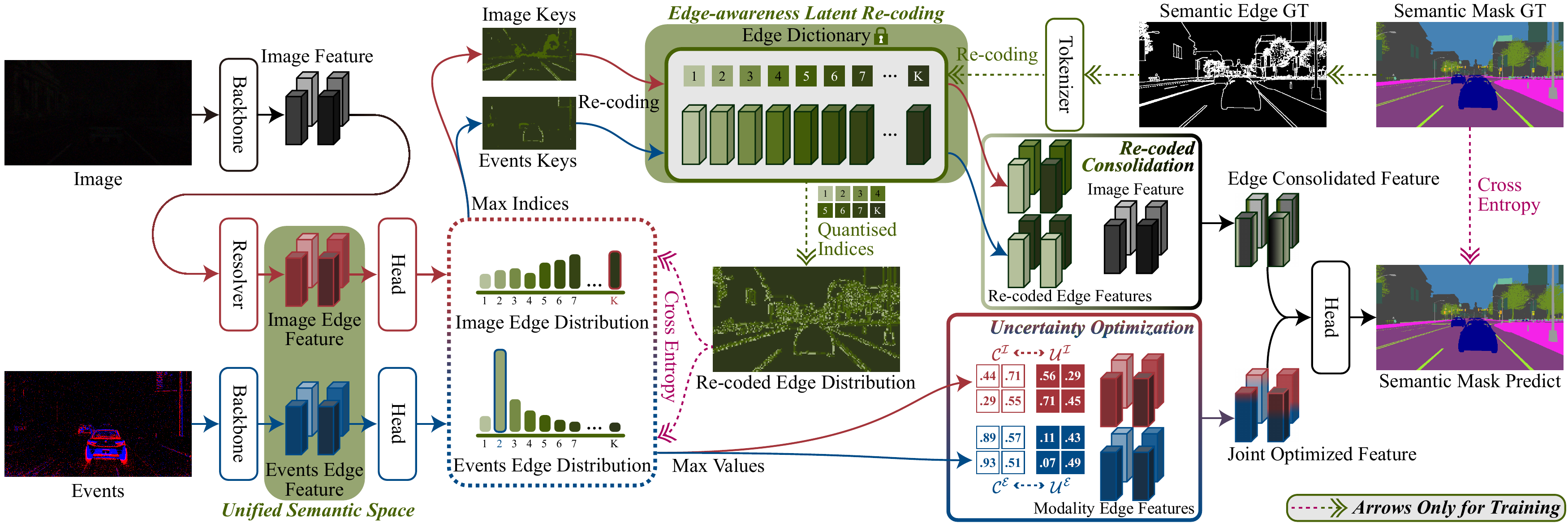}
    \captionsetup{font=small}
    \caption{
    \textbf{The overall architecture of our Edge-awareness Semantic Concordance (ESC). } ESC contains a pre-established edge dictionary and three key modules, namely Edge-awareness Latent Re-coding (ELR), Re-coded Consolidation (RC), and Uncertainty Optimization (UO). Based on the pre-trained edge dictionary, ELR transfers edge embeddings into re-coded distribution and modality distribution into re-coded features. RC consolidates edge information with re-coded features. UO jointly optimizes modality edge features with uncertainties. Features from RC and UO are concatenated for final semantic mask prediction. 
    }
    \label{fig:Overall}
\end{figure*}

\subsection{Cross-modality realignment of edge representations via latent re-coding}

\label{sec:ELR}

Re-coding is a key process in our framework, realigning edge representations of heterogeneous event and RGB through re-coded distribution, while also producing re-coded edge features for consolidation. 
Based on the pre-established edge dictionary, our proposed Edge-awareness Latent Re-coding module transfers edge embeddings into re-coded edge categorical prior distribution and modality posterior distribution into re-coded edge features.
This section will discuss the latent re-coding operation of two directions mentioned above, and introduce the optimization objective at the end of this section.

\textbf{Re-coding for edge categorical prior distribution.} 
Given the pre-trained tokenizer $f_T$ and the pre-established edge dictionary $\Delta$, we can re-code any semantic edge $\mathbf{B}$ to an edge categorical prior distribution $q(\mathcal{K}|\mathbf{B})$ as one-hot as follows: 
\begin{equation}
  \begin{split}
  q(\mathcal{K}|\mathbf{B}) = \mathop{\arg \min}_{\mathcal{K} \in \{\textbf{b}_k | k \in \{1, \cdots, K\}\}^{H'\times W'}}\|f_T(\mathbf{B}) - v(k)\|_2^2,
  \end{split}
  \label{eq:eq4-55}
\end{equation}
where $\textbf{b}_k$ is the $K$-dim basis vector with $1$ at $k$-th place.

\textbf{Re-coding for edge features.}
We first extract features from inputs. Given an image $\mathcal{I} \in \mathbb{R}^{H \times W \times 3}$ and its corresponding event voxel grid \cite{zihao2018unsupervised} $\mathcal{E} \in \mathbb{R}^{H \times W \times B}$, multi-scaled image feature and events feature can be obtained from $\mathbf{F}^{\mathcal{I}} = f_{\mathcal{I}}(\mathcal{I}), \mathbf{F}^{\mathcal{E}} = f_{\mathcal{E}}(\mathcal{E})$, where $B$ is number of voxel grid bins,  $\mathbf{F}^{\mathcal{I}}$ and $\mathbf{F}^{\mathcal{E}}$ denotes the backbones. Image edge features are resolved as $\mathbf{E}^{\mathcal{I}} = f_R(\mathbf{F}^{\mathcal{I}})$, where $f_R$ is an decouple module adopted from \cite{li2020improving} as our edge resolver. As event data naturally highlights the edge information, we keep event features directly as events edge features $\mathbf{E}^{\mathcal{E}} = \mathbf{F}^{\mathcal{E}}$ without additional processing. 
Then two edge encoders with the same structure are applied respectively to both $\mathbf{E}^{\mathcal{I}}$ and $\mathbf{E}^{\mathcal{E}}$, in order to encode modality edge features into the same unified semantic space.
Two MLP-based classification heads are attached after edge encoders for each modality to predict its modality-specific edge categorical probability distribution $p(\mathcal{K}|\mathcal{I})$ and $p(\mathcal{K}|\mathcal{E})$. This categorical probability distribution indicates the probability of edge dictionary item number $K$-ary classification at each spatial position.

Given probability distributions $p(\mathcal{K}|\mathcal{I})$ and $p(\mathcal{K}|\mathcal{E})$, we can retrieve image key map $\mathbf{K}^\mathcal{I}$ and events key map $\mathbf{K}^\mathcal{E}$ by
% $\mathcal{K}^{\mathcal{M}} = \mathop{\arg \max}_{k \in \{1, \cdots, K\}} p(\mathcal{K}=k|\mathcal{M}), \quad\mathcal{M} \in \{\mathcal{I}, \mathcal{E}\}$,
\begin{equation}
  \begin{split}
  \mathbf{K}^{\mathcal{M}} = \mathop{\arg \max}_{k \in \{1, \cdots, K\}} p(\mathcal{K}=k|\mathcal{M}), \quad\mathcal{M} \in \{\mathcal{I}, \mathcal{E}\},
  \end{split}
  \label{eq:eq4-7}
\end{equation}
where $\mathbf{K}^\mathcal{M} \in \{1, \cdots, K\}^{H' \times W'}$ means the key map of modality $\mathcal{M}$, which can either be image modality $\mathcal{I}$ or events modality $\mathcal{E}$. By this step, we select the indices of the maximum probability values at each position of the latent space as key map, which can be utilized to query edge dictionary for obtaining re-coded edge features of the specific modality. The image re-coded edge feature $\mathbf{\Gamma}^\mathcal{I}$ and events re-coded edge feature $\mathbf{\Gamma}^\mathcal{E}$ are obtained by 
\begin{equation}
  \begin{split}
  \mathbf{\Gamma}^\mathcal{M} = v(\mathbf{K}^\mathcal{M}), \quad\mathcal{M} \in \{\mathcal{I}, \mathcal{E}\},
  \end{split}
  \label{eq:eq4-8}
\end{equation}
where $\mathbf{\Gamma}^\mathcal{M} \in \mathbb{R}^{H' \times W' \times n}$ are modality-specific edge embeddings queried by the specific key map.

\textbf{How to optimize ELR and what are the benefits?}
We optimize our Edge-awareness Latent Re-coding module by an objective function based on cross-entropy, which narrows the gap between the edge categorical distribution $q(\mathcal{K}|\mathbf{B})$ with modality-specific edge categorical probability distribution $p(\mathcal{K}|\mathcal{I})$ and $p(\mathcal{K}|\mathcal{E})$ as $L_{edge} = - \sum q(\mathcal{K}|\mathbf{B}) \log (p(\mathcal{K}|\mathcal{I}) p(\mathcal{K}|\mathcal{E})),$ a summation of two cross-entropies.
By minimizing this objective function, we can bridge the modality gap and realign the image edge feature $\mathbf{E}^{\mathcal{I}}$ with events edge feature $\mathbf{E}^{\mathcal{E}}$ into the same unified semantic space, and make sure the re-coded features $\mathbf{\Gamma}^{\mathcal{M}}$ represent the edge information of events and image correctly. 

\subsection{Re-coded features for edge information consolidation}

\label{sec:REC}

% NIPS CR Ver.
\begin{wrapfigure}{r}{0.6\textwidth}
    \centering
    \vspace{-0.5em}
    \includegraphics[width=1.0\linewidth]{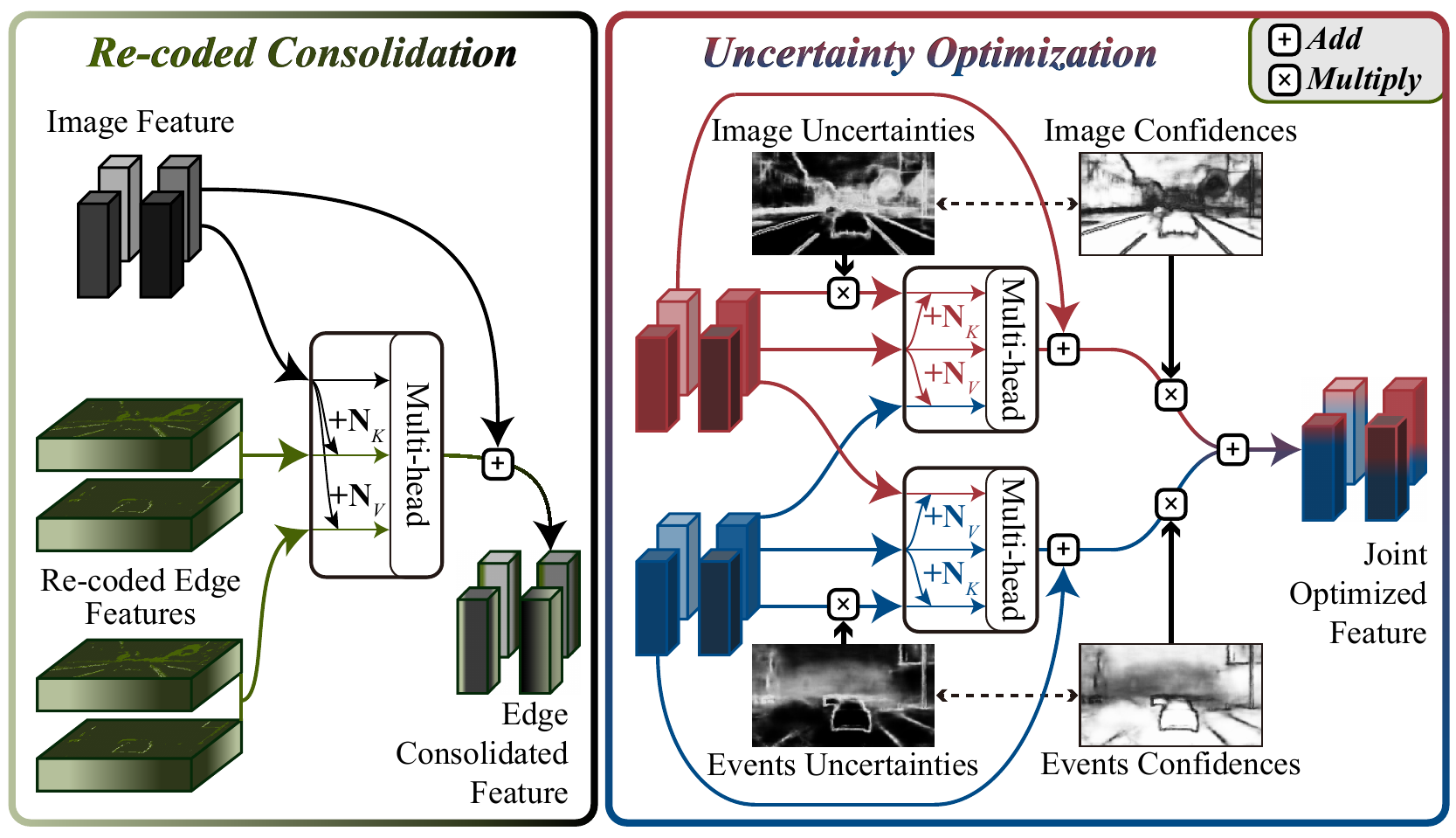}
    \captionsetup{font=small}
    \caption{\textbf{RC and UO.} The two modules utilize an attention-based structure with learnable noise embeddings for a resilient fusion.}
    \label{fig:SubArch}
    \vspace{-1em}
\end{wrapfigure}

Image feature mainly focuses on contextual information and lack of understanding of edge information. Re-coded features are utilized for edge information consolidation by the Re-coded Consolidation (RC) module. As shown in \cref{fig:SubArch} on the left, RC takes image feature $\mathbf{F}^\mathcal{I}$, image and events re-coded edge feature $\mathbf{\Gamma}^\mathcal{I}$ and $\mathbf{\Gamma}^\mathcal{E}$ as input, outputs a refined feature namely Edge Consolidated Feature $\mathbf{\Phi}$.

In RC, we define two learnable noise embeddings $\mathbf{N}_K \in \mathbb{R}^n$ and $\mathbf{N}_V \in \mathbb{R}^n$ to improve fitting ability and enhance learning stability. For image feature $\mathbf{F}^\mathcal{I}$ and re-coded edge features $\mathbf{\Gamma}^\mathcal{I}$ and $\mathbf{\Gamma}^\mathcal{E}$, RC applies multi-head attention operation on vectors at spatial position $\langle h, w \rangle$, calculates and outputs the consolidated features $\mathbf{\Phi}$ as
\begin{equation}
  \begin{split}
  \mathbf{\Phi}_{h, w} &= [\phi_1, \phi_2, \cdots, \phi_m] \cdot W_O + \mathbf{F}^\mathcal{I}_{h, w}, \\
  \phi_i &= \mbox{Softmax}(Q_iK_i^{\mathsf{T}}/\sqrt{d_k}) \cdot 
 V_i,
  \label{eq:eq4-9}
  \end{split}
\end{equation}
where $Q_i = \mathbf{F}^\mathcal{I}_{h, w} \cdot W_{Q_i}$, $K_i = [\mathbf{F}^\mathcal{I}_{h, w} + \mathbf{N}_K, \mathbf{\Gamma}^\mathcal{I}_{h, w}, \mathbf{\Gamma}^\mathcal{E}_{h, w}] \cdot W_{K_i}$, $V_i = [\mathbf{F}^\mathcal{I}_{h, w} + \mathbf{N}_V, \mathbf{\Gamma}^\mathcal{I}_{h, w}, \mathbf{\Gamma}^\mathcal{E}_{h, w}] \cdot W_{V_i}$.
The introduction of noise embedding is inspired by \cite{DBLP:conf/icml/JiaG0W00024}, and we develop the following theoretical explanation.
In the absence of noise, the query ($Q_i$) tends to attend excessively to its own features in the key ($K_i$), thereby suppressing signals from the other source and impeding effective fusion. Introducing noise embeddings mitigates this issue by perturbing the attention space in a controlled, learnable manner, encouraging richer and more balanced cross-modality interactions.
The main idea of \cref{eq:eq4-9} is to consolidate image feature $\mathbf{F}^\mathcal{I}$ with image re-coded edge feature $\mathbf{\Gamma}^\mathcal{I}$ and events re-coded edge feature $\mathbf{\Gamma}^\mathcal{E}$ by querying $\mathbf{\Gamma}^\mathcal{I}$ and $\mathbf{\Gamma}^\mathcal{E}$ with $\mathbf{F}^\mathcal{I}$ to obtain attention map. The map represents the relevance between $\mathbf{F}^\mathcal{I}$ and $\mathbf{\Gamma}^\mathcal{I}, \mathbf{\Gamma}^\mathcal{E}$, which decides the amount of edge information consolidated by $\mathbf{F}^\mathcal{I}$. Refined vectors $\mathbf{\Phi}_{h, w}$ constitute together in accordance with their positions as $\mathbf{\Phi} \in \mathbf{R}^{H_d \times W_d \times n}$.

\subsection{Edge-aware uncertainties for joint optimization}

\label{sec:UEJO}

Probability values in edge distribution indicate the confident and uncertain areas of image and events. We leverage this confidence and uncertainty information from edge distribution of image and events for a resilient fusion by the Uncertainty Optimization (UO) module.

Given edge categorical probability distributions $p(\mathcal{K}|\mathcal{I})$ and $p(\mathcal{K}|\mathcal{E})$, we can retrieve confidences and uncertainties of image and events by
\begin{equation}
  \begin{split}
  \mathcal{C}^{\mathcal{M}} &= \max_{k \in \{1, \cdots, K\}} p(\mathcal{K}=k|\mathcal{M}), \\
  \mathcal{U}^{\mathcal{M}} &= 1 - \mathcal{C}^{\mathcal{M}}, \quad \mathcal{M} \in \{\mathcal{I}, \mathcal{E}\},
  \end{split}
  \label{eq:eq4-10}
\end{equation}
where $\mathcal{C}^{\mathcal{M}}, \mathcal{U}^{\mathcal{M}} \in [0, 1]^{H' \times W'}$ denote the confidences and uncertainties of modality $\mathcal{M}$, which can either be image modality $\mathcal{I}$ or events modality $\mathcal{E}$. 

Confidences and Uncertainties represent spatial reliability of specific modality, which are utilized as indicators in UO. As shown in \cref{fig:SubArch} on the right, UO takes image edge feature $\mathbf{E}^{\mathcal{I}}$ and events edge feature $\mathbf{E}^{\mathcal{E}}$ as input, confidences and uncertainties as indicators, and outputs a refined feature namely Joint Optimized Feature $\mathbf{\Psi}$.

In UO, we define four learnable noise embeddings $\mathbf{N}_K^\mathcal{I}, \mathbf{N}_K^\mathcal{E}, \mathbf{N}_V^\mathcal{I}, \mathbf{N}_V^\mathcal{E} \in \mathbb{R}^n$ to enhance fitting capability and improve learning robustness. For feature $\mathbf{E}^{\mathcal{I}}$ and $\mathbf{E}^{\mathcal{E}}$, their confidences and uncertainties are $\mathcal{C}^\mathcal{I}$, $\mathcal{U}^\mathcal{I}$ and $\mathcal{C}^\mathcal{E}$, $\mathcal{U}^\mathcal{E}$ respectively. UO applies multi-head attention operation on vectors at spatial position $\langle h, w \rangle$, calculates and outputs the optimized feature $\mathbf{\Psi}$ as
\begin{equation}
  \begin{split}
  \mathbf{\Psi}_{h, w} &= 
  \frac{\mathcal{C}^\mathcal{I}_{h, w} \cdot \mathbf{\Psi}^{\mathcal{I}}_{h, w}}{\mathcal{C}^\mathcal{I}_{h, w} + \mathcal{C}^\mathcal{E}_{h, w}} + 
  \frac{\mathcal{C}^\mathcal{E}_{h, w} \cdot \mathbf{\Psi}^{\mathcal{E}}_{h, w}}{\mathcal{C}^\mathcal{I}_{h, w} + \mathcal{C}^\mathcal{E}_{h, w}}, \\
  % Image Attention
  \mathbf{\Psi}^{\mathcal{M}}_{h, w} &= 
  [\psi^{\mathcal{M}}_1, \psi^{\mathcal{M}}_2, \cdots, \psi^{\mathcal{M}}_m] \cdot W^\mathcal{M}_O + \mathbf{E}^{\mathcal{M}}_{h, w}, \\ 
  \psi^{\mathcal{M}}_i &= 
  \mbox{Softmax}(Q^\mathcal{M}_i (K^\mathcal{M}_i)^{\mathsf{T}} / \sqrt{d_k}) \cdot V^\mathcal{M}_i,
  \end{split}
  \label{eq:eq4-11}
\end{equation}
where $Q^\mathcal{M}_i=\mathbf{E}^{\mathcal{M}}_{h, w} \cdot W_{Q_i}^\mathcal{M}$, $K^\mathcal{M}_i = [\mathbf{E}^{\mathcal{M}}_{h, w} + \mathbf{N}_K^\mathcal{M}, \mathcal{U}^\mathcal{M}_{h, w} \cdot\mathbf{E}^{\mathcal{M}}_{h, w}] \cdot W_{K_i}^\mathcal{M}$, $V^\mathcal{M}_i = [\mathbf{E}^{\mathcal{M}}_{h, w} + \mathbf{N}_V^\mathcal{M}$, $\mathbf{E}^{\overline{\mathcal{M}}}_{h, w}] \cdot W_{V_i}^\mathcal{M}$,
  $\langle \mathcal{M}, \overline{\mathcal{M}} \rangle \in \{\langle \mathcal{I}, \mathcal{E}\rangle, \langle \mathcal{E}, \mathcal{I}\rangle\}$.
The main idea of \cref{eq:eq4-11} is to optimize image edge feature $\mathbf{E}^{\mathcal{I}}$ and events edge feature $\mathbf{E}^{\mathcal{E}}$ based on their confidences at each spatial position. Multiplied by the uncertainty value $\mathcal{U}^\mathcal{M}$, modality-specific feature exposes its uncertainty to attention map. The map represents the self-uncertainty of $\mathbf{E}^{\mathcal{I}}$ and $\mathbf{E}^{\mathcal{E}}$, which decides the amount of complementary information absorbed from the counter modality. The final feature vector is calculated by normalized confidence weighted summation of inter-modalities feature vectors. Optimized vectors $\mathbf{\Psi}_{h, w}$ constitute together in accordance with their positions as $\mathbf{\Psi} \in \mathbf{R}^{H_d \times W_d \times n}$.

\textbf{How to optimize our method?}
Edge consolidate feature $\mathbf{\Phi}$ and joint optimized feature $\mathbf{\Psi}$ are concatenated and input into an MLP-based classification head for semantic mask prediction. Cross-entropy is utilized for the supervision of semantic mask prediction as $L_{pred}$. The final optimization objective function for our method is $L = L_{pred}+ \beta \cdot L_{edge}$, where $\beta$ is a constant of edge loss weight.

\subsection{Constructing datasets for reliable evaluation of event-RGB segmentation}

\label{sec:datasets}

\begin{figure*}[t!]
    \centering
    \includegraphics[width=1.0\linewidth]{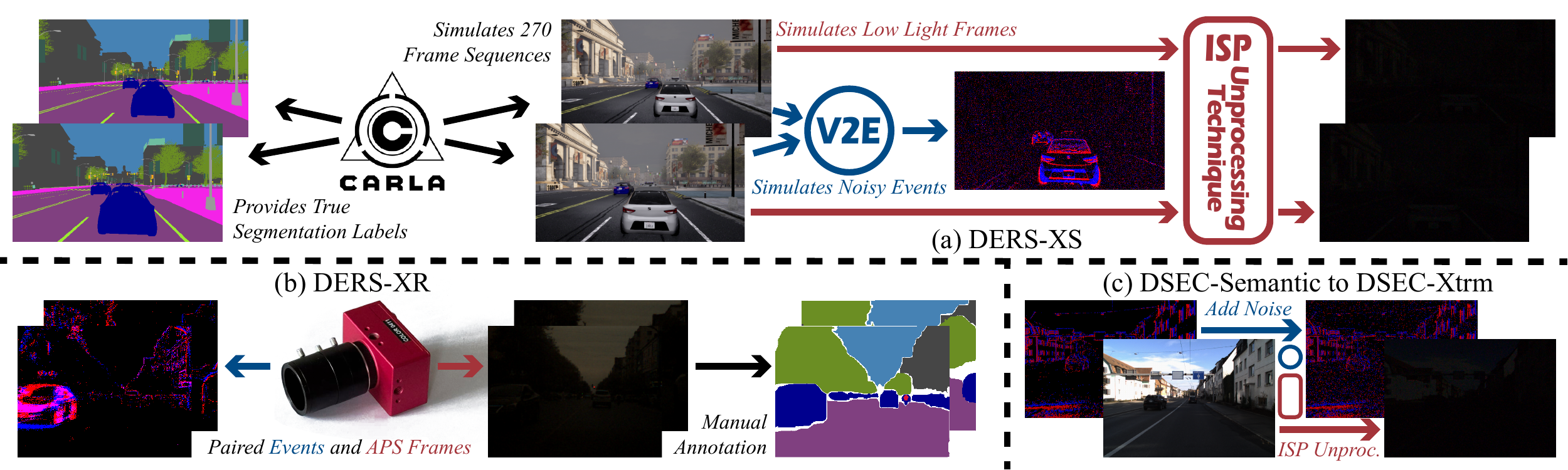}
    \captionsetup{font=small}
    \caption{Construction of datasets DERS-XS, DERS-XR, and DSEC-Xtrm for reliable evaluation.}
    \label{fig:DATASETS}
\end{figure*}

Labels from DSEC-Semantic are pseudo-labels directly derived from RGB via \cite{tao2020hierarchical}. They are useful for event-only tasks, but not reliable for evaluating event-RGB tasks. Using these labels as ground truth implicitly presupposes that the optimal result of event-RGB segmentation is obtained by an RGB-only model, which undermines the unique advantages of events. To address the problem, we construct datasets as below for reliable evaluation.
The construction pipelines of datasets DERS-XS, DERS-XR, and DSEC-Xtrm are shown in \cref{fig:DATASETS}.
Details of the datasets can be seen in \cref{sec:appendixA}.

\textbf{DERS-XS}. Dataset of Event-RGB semantic Segmentation under eXtreme conditions Synthetic, abbreviated as DERS-XS, is a true-labeled synthetic event-RGB extreme condition semantic segmentation dataset. \textbf{CARLA} \cite{Dosovitskiy17} \textbf{provides true segmentation labels}, and we first use CARLA simulator to simulate 270 frame sequences with segmentation labels of 23 categories, each with 1200 frames, and the size of each frame is 640 \texttimes~360. We then simulate noisy events from the frame sequences using v2e simulator \cite{hu2021v2e} with a shot noise parameter of 5.0 Hz. Low-light frames are simulated by attenuating optical signals and adding shot noise in the RAW domain obtained from the ISP unprocessing technique in \cite{brooks2019unprocessing}.
Because differences between adjacent frames are small, which is not conducive to data diversity, we sample data at intervals of 100 frames while discarding other frames. We divide 168 sequences as training set, 12 sequences as validation set, and 90 sequences as test set. 

\textbf{DERS-XR}. Dataset of Event-RGB semantic Segmentation under eXtreme conditions Real-world, abbreviated as DERS-XR, is a \textbf{manually annotated} real-world event-RGB extreme condition semantic segmentation dataset. We use a DAVIS346 \cite{taverni2018front} to capture paired APS frames and events under extreme lighting conditions, and manually annotate a subset of 240 frames. Of these, 120 frames are randomly selected for fine-tuning, while the remaining 120 frames are used for testing.

\textbf{DSEC-Xtrm}. DSEC-Xtrm is an extreme condition semantic segmentation dataset synthesized based on DSEC-Semantic \cite{gehrig2021dsec, sun2022ess}.
\textbf{To make use of pseudo-labels from DSEC-Semantic while mitigating their direct dependence on RGB}, we 
apply degradation to the RGB frames.
We apply the same low-light image simulation method as DERS-XS to frames and use v2e simulator \cite{hu2021v2e} to generate pure shot noise and add it to events. The degraded frames and events together constitute DSEC-Xtrm.
\section{Experiments}

\subsection{Implement details}

\label{sec:4.1}

The code is implemented by PyTorch. We first train edge dictionary as a separate stage to obtain pre-trained weights of tokenizer and edge dictionary. We utilize pre-trained MiT-B2 backbone and MiT-B1 backbone of SegFormer \cite{xie2021segformer} for RGB modality and event modality, respectively. The number of categories $c$ is 11. We set the number of items $K$ in edge dictionary as 128, and the dimension of edge embeddings $n$ as 256. The weight of edge dictionary commitment loss $\alpha$ is 0.25, and the weight of edge loss $\beta$ is 0.1. The bins of event voxel grid $B$ is 5. For training, we randomly apply color jitter, horizontal flipping, and gaussian blur to images and randomly resize with scales from 0.5 to 2.0 to images and events and crop the inputs to 256 \texttimes~256. For testing, we follow the setting of CMX \cite{zhang2023cmx} and CMNeXt \cite{zhang2023delivering}, which upsamples the inputs to a width and height both divisible by 32 (will be ablated in \cref{tab:abla_abla}). More detailed information on training settings can be seen in \cref{sec:appendixB}.

\subsection{Comparisons with state-of-the-art}

\begin{table}[t]
\centering
\captionsetup{font=small}
\caption{Comparisons on DERS-XS, DERS-XR, DSEC-Semantic, and DSEC-Xtrm.}\label{tab:comp_all}
\setlength\tabcolsep{1pt}
\renewcommand{\arraystretch}{1}
\resizebox{1\textwidth}{!}{
\begin{threeparttable}[t]
\begin{tabular}{l c c c c c c c c c c c}
\mytoprule
 \multirow{2}{*}{Methods} & \multirow{2}{*}{Modality} & \multicolumn{2}{c}{DERS-XS} & \multicolumn{2}{|c}{DERS-XR} & \multicolumn{2}{|c}{DSEC-Semantic} & \multicolumn{2}{|c}{DSEC-Xtrm}  \\ 
 \mycmidrule{3-10}
  &  & mACC(\%)\textuparrow & mIoU(\%)\textuparrow & mACC(\%)\textuparrow & mIoU(\%)\textuparrow & mACC(\%)\textuparrow & mIoU(\%)\textuparrow & mACC(\%)\textuparrow & mIoU(\%)\textuparrow \\ 
 \mymidrule
 SegFormer \cite{xie2021segformer} & RGB & 62.45 & 53.21 & 55.37 & 51.09 & 72.91 & 65.03 & 41.74 & 33.88 \\ 
 \mymidrule
 SegFormer (E) \cite{xie2021segformer} & Event & 47.85 & 37.32 & 42.03 & 36.96 & 47.38 & 38.59 & 48.52 & 37.72 \\
 EvSegFormer \cite{jia2023event} & Event & 41.48 & 31.85 & 38.60 & 33.66 & 44.72 & 37.13 & 42.33 & 34.68 \\
 \mymidrule
 TokenFusion \cite{wang2022multimodal} & E-RGB & 64.88 & 56.22 & 54.19 & 47.72 & 74.60 & 67.39 & \cellcolor{third}53.04 & \cellcolor{third}45.41 \\
 CMX \cite{zhang2023cmx} & E-RGB & \cellcolor{third}71.86 & \cellcolor{third}63.12 & 64.51 & 59.22 & \cellcolor{third}76.18 & \cellcolor{third}68.10 & 51.29 & 43.95 \\
 CMNeXt \cite{zhang2023delivering} & E-RGB & \cellcolor{second}73.30 & \cellcolor{second}64.55 & \cellcolor{second}66.57 & \cellcolor{third}60.95 & \cellcolor{second}77.50 & \cellcolor{second}69.03 & 52.12 & 45.16 \\
 EISNet \cite{xie2024eisnet} \tnote{\dag} & E-RGB & 69.10 & 60.68 & \cellcolor{third}66.18 & \cellcolor{second}61.81 & 71.60 & 64.67 & \cellcolor{second}56.77 & \cellcolor{second}48.76 \\
 \textbf{Ours} & E-RGB & \cellcolor{first}\textbf{75.26} & \cellcolor{first}\textbf{67.10} & \cellcolor{first}\textbf{70.75} & \cellcolor{first}\textbf{65.22} & \cellcolor{first}\textbf{78.61} & \cellcolor{first}\textbf{71.04} & \cellcolor{first}\textbf{59.45} & \cellcolor{first}\textbf{50.87} \\
 \mybottomrule
\end{tabular}
\begin{tablenotes}
 \item[\dag] Reimplemented on DSEC-Semantic using the same dataloader for fair comparison with different training settings from \cite{xie2024eisnet}, including each sequence events count of 50000 (\cite{xie2024eisnet} of 100000), input cropping size of 256 \texttimes~256 (\cite{xie2024eisnet} of 448 \texttimes~448), total batch size of 32 (\cite{xie2024eisnet} of 8), different random resize strategy, \etc.
\end{tablenotes}
\end{threeparttable}
}
\end{table}

\begin{figure*}[t!]
    \centering
    \includegraphics[width=1.0\linewidth]{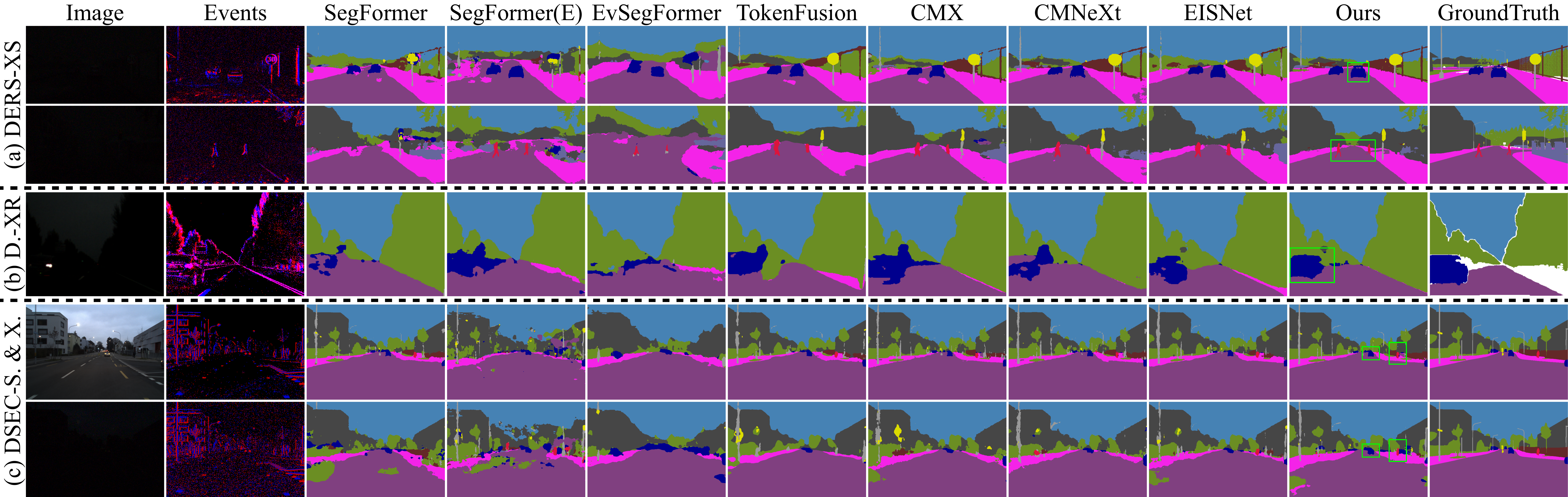}
    % \vspace{-1.5em}
    \captionsetup{font=small}
    \caption{Qualitative comparison on DERS-XS, DERS-XR, DSEC-Semantic, and DSEC-Xtrm.}
    \label{fig:VIZ-ALL}
\end{figure*}

We compare our model with the current state-of-the-art, including RGB-only, event-only, and event-RGB-based methods. For RGB-only, we reimplement SegFormer \cite{xie2021segformer} with MiT-B2 backbone settings, which is a powerful RGB-based semantic segmentation architecture. For event-only, we reimplement EvSegFormer \cite{jia2023event} and modify the number of input channels of SegFormer to adapt to events input as an events-only comparison method. For event-RGB-based, we reimplement TokenFusion \cite{wang2022multimodal}, CMX \cite{zhang2023cmx}, CMNeXt \cite{zhang2023delivering}, and EISNet \cite{DBLP:conf/iccv/XiaZZWST23}. For event-version SegFormer and EvSegFormer, we follow the setting of EvSegFormer and apply 6-channel image \cite{alonso2019ev} as their event representation. For TokenFusion, CMX, CMNeXt, and EISNet, we use 3-bin voxel grid \cite{zihao2018unsupervised} as their event representation. \textbf{We retrain all methods with the same training settings for fair comparison.}

\textbf{Comparisons on true-labeled synthetic dataset.} As shown in \cref{tab:comp_all}, our method surpasses all uni-modality and multi-modality methods and outperforms CMNeXt by a 2.55\% mIoU on DERS-XS. As shown in \cref{fig:VIZ-ALL}a, our method is more stable and robust to the edges of the segmentation results, especially for moving vehicles and pedestrians. This shows that our model effectively leverages edge information from events to compensate for the information loss of RGB under extreme conditions.

\textbf{Comparisons on true-labeled real-world dataset fine-tuning.} As shown in \cref{tab:comp_all}, our methods outperforms EISNet by a 3.41\% mIoU on DERS-XR fine-tuning experiments. Results also demonstrate that models trained on synthetic DERS-XS can be efficiently adapted to real-world data with minimal fine-tuning, further validating the effectiveness of DERS-XS. As shown in \cref{fig:VIZ-ALL}b, our method successfully segments the vehicles, while other methods fail under real-world extreme scenes.

\textbf{Comparisons on non-extreme DSEC-Semantic.} As shown in \cref{tab:comp_all}, our method outperforms CMNeXt by a 2.01\% mIoU on DSEC-Semantic, demonstrating its effectiveness on a publicly available dataset under real-world, non-extreme conditions, despite the inherent limitations of pseudo-labels. EISNet performs slightly worse, possibly due to its sensitivity to the input cropping strategy.

\textbf{Comparisons under extreme conditions on degraded DSEC-Xtrm.} As shown in \cref{tab:comp_all}, our method ourperforms EISNet by a 2.11\% mIoU on DSEC-Xtrm. 
Results also show that our method suffers less performance degradation with degraded inputs. As shown in \cref{fig:VIZ-ALL}c, our method preserves more complete boundaries for vehicles and pedestrians, demonstrating its robustness and resilience.

\subsection{Ablation studies and analyses}

\begin{wraptable}{r}{0.5\textwidth}
\vspace{-1.5em}
\centering
\captionsetup{font=small}

\caption{Study on spatial occlusion on DERS-XS.}\label{tab:comp_mask}
\vspace{-0.5em}
\setlength{\tabcolsep}{8pt}
\renewcommand{\arraystretch}{1}
\resizebox{1\linewidth}{!}{
\begin{tabular}{l c c c c}
\mytoprule
 \multirow{2}{*}{Methods\,/\,mIoU(\%)\textuparrow} & \multicolumn{4}{c}{Apply masking on} \\ 
 \mycmidrule{2-5}
 & None & RGB & Event & E-RGB\\ 
 \mymidrule
 TokenFusion \cite{wang2022multimodal} & 56.22 & 48.44 & 55.79 & 48.00 \\
 CMX \cite{zhang2023cmx} & \cellcolor{third}63.12 & 54.13 & \cellcolor{third}62.70 & \cellcolor{third}53.73 \\
 CMNeXt \cite{zhang2023delivering} & \cellcolor{second}64.55 & \cellcolor{third}54.15 & \cellcolor{second}64.07 & 53.70 \\
 EISNet \cite{xie2024eisnet} & 60.68 & \cellcolor{second}55.33 & 59.87 & \cellcolor{second}54.47 \\
 % \midrule
 \textbf{Ours} & \cellcolor{first}\textbf{67.10} & \cellcolor{first}\textbf{64.34} & \cellcolor{first}\textbf{66.65} & \cellcolor{first}\textbf{63.87} \\
 \mybottomrule
\end{tabular}
}

\vspace{0.5em}
\caption{Ablation study on architecture on DERS-XS.}\label{tab:abla_abla}
\vspace{-0.5em}
\setlength{\tabcolsep}{4pt}
\renewcommand{\arraystretch}{1}
\resizebox{1\linewidth}{!}{
\begin{tabular}{l|l|l|l}
\mytoprule
 Arch.\,/\,mIoU(\%)\textuparrow & \#Params(M) & w/o masking & E-RGB masking \\ 
 \mydmidrule
 ESC & 56.875 & 67.10 & 63.87 \\
 % \midrule
 - w/o Upsampling & 56.875 & 64.59 & 61.46 \\
 \mymidrule
 - w/o RC & 56.612 & 64.29 \textcolor{G28}{\small{(-0.31)}} & 59.53 \textcolor{G28}{\small{(-1.93)}} \\
 - w/o UO & 56.084 & 62.53 \textcolor{G28}{\small{(-2.06)}} & 58.43 \textcolor{G28}{\small{(-3.03)}} \\
 - w/o ELR\&$L_{edge}$ & 38.411 & 61.35 \textcolor{G28}{\small{(-3.24)}} & 56.34 \textcolor{G28}{\small{(-5.12)}} \\
 \mybottomrule
\end{tabular}
}

\vspace{0.5em}

\caption{Ablation study on key usage on DERS-XS.}\label{tab:abla_keyn}
\vspace{-0.5em}
\setlength\tabcolsep{4pt}
\renewcommand{\arraystretch}{1}
\resizebox{1\linewidth}{!}{
\begin{threeparttable}[t]
\begin{tabular}{l c c c c c c}
\mytoprule
 K & 16 & 32 & 64 & 128 & 256 & 512 \\ 
 \mymidrule
 K-Usage\tnote{\dag} & 16 & 32 & 64 & 92 & 99 & 97 \\
 gACC(\%)\textuparrow & 92.66 & \cellcolor{third}93.12 & 93.03 & \cellcolor{first}\textbf{93.27} & \cellcolor{second}93.19 & 92.93 \\
 mACC(\%)\textuparrow & 74.59 & \cellcolor{third}74.77 & \cellcolor{second}74.90 & \cellcolor{first}\textbf{75.26} & 73.91 & 73.65 \\
 mIoU(\%)\textuparrow & 65.87 & \cellcolor{second}66.84 & \cellcolor{third}66.64 & \cellcolor{first}\textbf{67.10} & 66.54 & 66.11 \\
 \mybottomrule
\end{tabular}
\begin{tablenotes}
     \item[\dag]: K-Usage is the number of dictionary keys used. 
\end{tablenotes}
\end{threeparttable}
}

\vspace{0.5em}

\caption{Ablation study on noise embeddings removal.}\label{tab:embeddings}
\vspace{-0.5em}
\setlength\tabcolsep{2pt}
\renewcommand{\arraystretch}{1}
\resizebox{1\linewidth}{!}{
\begin{tabular}{l c c c}
\mytoprule
 Arch.\,/\,mIoU(\%)\textuparrow & DSEC-XS & DSEC-Semantic & DSEC-Xtrm \\ 
 \mymidrule
 ESC w/o $\textbf{N}_{K}, \textbf{N}_{V}$ & 66.05 & 70.86 & 50.05 \\
 ESC w/ $\textbf{N}_{K}, \textbf{N}_{V}$ & 67.10 & 71.04 & 50.87 \\
 \mybottomrule
\end{tabular}
}

\vspace{0.5em}

\caption{Comp. on model complexity on DERS-XS.}\label{tab:comp_comp}
\vspace{-0.5em}
\setlength{\tabcolsep}{4pt}
\renewcommand{\arraystretch}{1}
\resizebox{1\linewidth}{!}{
\begin{tabular}{l c c c c}
\mytoprule
 Methods &  Backbone & \#Params(M) & FLOPs(G) & mIoU(\%)\textuparrow\\ 
 \mymidrule
 SegFormer \cite{xie2021segformer} & MiT-B2 & 24.725 & 25.279 & 53.21 \\ 
 \mymidrule
 SegFormer (E) \cite{xie2021segformer} & MiT-B2 & 24.734 & 25.433 & 37.32 \\
 EvSegFormer \cite{jia2023event} & MiT-B2 & 24.740 & 25.422 & 31.85 \\
 \mymidrule
 TokenFusion \cite{wang2022multimodal} & MiT-B2 & 26.011 & 54.845 & 56.22 \\
 CMX \cite{zhang2023cmx} & 2 \texttimes~MiT-B2 & 66.566 & 65.551 & \cellcolor{third}63.12 \\
 CMNeXt \cite{zhang2023delivering} & 2 \texttimes~MiT-B2 & 58.687 & 62.805 & \cellcolor{second}64.55 \\
 EISNet \cite{xie2024eisnet} & MiT-B0 + B2 & 34.367 & 67.304 & 60.68 \\
 \textbf{Ours} & MiT-B1 + B2 & 56.875 & 95.086 & \cellcolor{first}\textbf{67.10} \\
 \mybottomrule
\end{tabular}
}

\vspace{-1.5em}

\end{wraptable}

\begin{wrapfigure}{r}{0.5\textwidth}
    \centering
    \vspace{-1.25em}
    \includegraphics[width=1.0\linewidth]{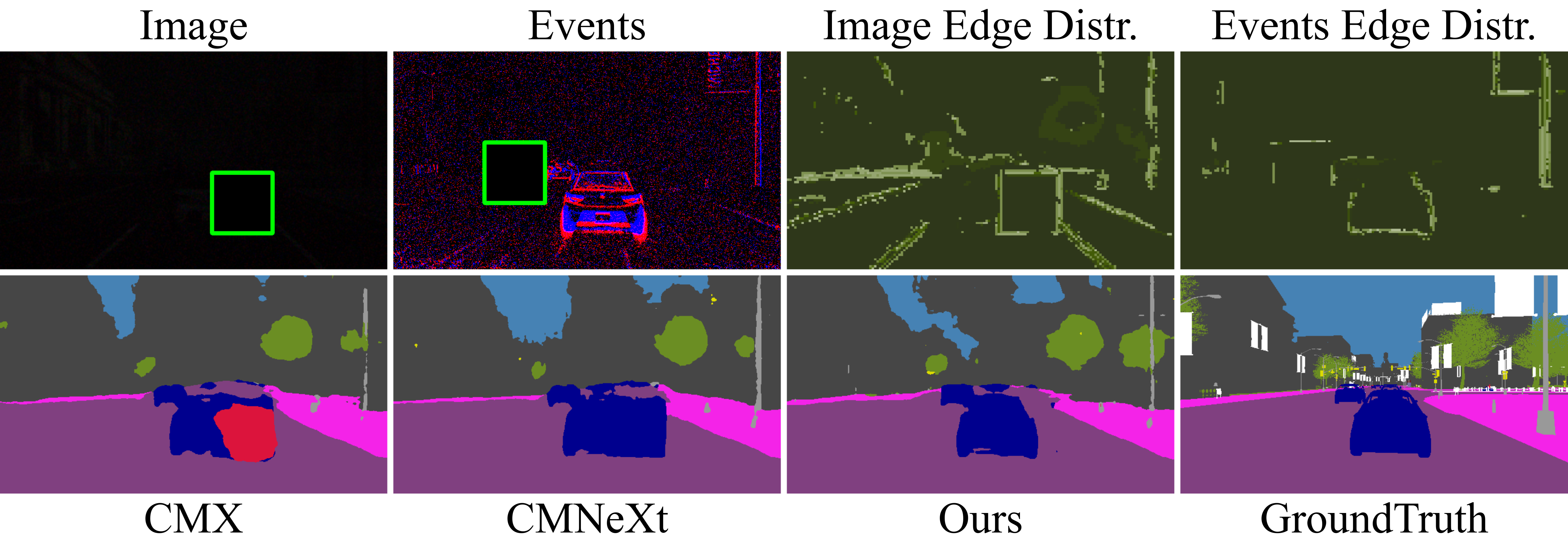}
    \captionsetup{font=small}
    \caption{Qualitative study under spatial occlusion.}
    \label{fig:Mask_VIZ}
    \vspace{-1em}
\end{wrapfigure}

\textbf{Resilience study under severe spatial occlusion.} This study emulates visual degradation due to spatial information loss under extreme conditions by applying local masking to the inputs. 
If masking is applied, we mask a 100 \texttimes~100 area starting at coordinates \textlangle350, 200\textrangle~for RGB and \textlangle150, 150\textrangle~for event.
As shown in \cref{tab:comp_mask}, our method suffers less performance degradation under different settings.
As shown in \cref{fig:Mask_VIZ}, under E-RGB masking, CMX and CMNeXt fail in understanding the semantics of the mask area, while our method overcomes the problem by edge-aware optimization with uncertainty indicators.
This demonstrates that our method is more resilient than other methods under severe spatial occlusion.
Extended experiments with more results can be seen in \cref{sec:appendixE} and \ref{sec:appendixF}.

\textbf{Ablation study on ESC architecture.} As shown in \cref{tab:abla_abla}, we ablate our ESC by removing modules under no mask setting and E-RGB mask setting on DERS-XS. 
Upsampling, as a form of data augmentation, is first ablated. 
Results show that each proposed module contributes positively to performance and resilience.

\textbf{Ablation study on different $K$ of edge dictionary.}
As shown in \cref{tab:abla_keyn}, when $K$ is too small, the items are insufficient for edge representation; when $K$ is too large, the excess items are underutilized, 
leading to confusion in model learning.
Both result in performance degradation; thus, we set $K=128$ to achieve an optimal trade-off.

\textbf{Ablation study on the removal of noise embeddings.} As shown in \cref{tab:embeddings}, the removal of noise embeddings leads to a 1.05\% and 0.81\% mIoU drop on DERS-XS and DSEC-Xtrm, respectively, confirming their contribution to improved fitting stability. The performance drop on DSEC-Semantic is minimal (0.18\%), which we attribute to its reliance on the RGB-based pseudo-labels. As the supervision signal has a bias towards RGB, the model naturally relies less on event modality. In such cases, the role of noise embeddings in facilitating cross-modality interaction becomes less significant.

\textbf{Model complexity.} \Cref{tab:comp_comp} summarizes the complexity of compared models, where FLOPs are calculated with inputs of 512 \texttimes~512. Results show that our method has fewer parameters than CMX and CMNeXt, yet achieves better performance.
The FLOPs of our method are relatively large, primarily due to multiple MLP heads for re-coding and resilient fusion in our framework.
\section{Conclusions and limitations}

\label{sec:5}

\textbf{Conclusions.} In this paper, we propose Edge-awareness Semantic Concordance, a multi-modality framework for event-RGB semantic segmentation. We demonstrate its capability and robustness for handling heterogeneous event and RGB. Results show that our framework outperforms existing event-RGB segmentation methods and possesses superior resilience in the case of modality imbalance and failure under extreme conditions. 
\textbf{Limitations.} Despite the promising results, only the fusion of event and RGB is considered so far. Exploring interactions with other visual modalities and designing modules tailored to their specific characteristics continues to be an open direction for future research.

\clearpage

\section*{Acknowledgements}

This work is partially supported by grants from the National Natural Science Foundation of China under contracts No. 62132002 and No. 62202010, the Beijing Nova Program (No.20250484786), and the Fundamental Research Funds for the Central Universities.

{
    \small
    \bibliographystyle{plain}
    \bibliography{ref}

@String(ICCV= {Int. Conf. Comput. Vis.})

@String(ECCV= {Eur. Conf. Comput. Vis.})

@String(ICLR = {Int. Conf. Learn. Represent.})

@String(AAAI = {AAAI})

@String(ICCV  = {ICCV})

@String(ECCV  = {ECCV})

@String(ICLR  = {ICLR})

@inproceedings{li2020improving,
  title={Improving semantic segmentation via decoupled body and edge supervision},
  author={Li, Xiangtai and Li, Xia and Zhang, Li and Cheng, Guangliang and Shi, Jianping and Lin, Zhouchen and Tan, Shaohua and Tong, Yunhai},
  booktitle={Computer Vision--ECCV 2020: 16th European Conference, Glasgow, UK, August 23--28, 2020, Proceedings, Part XVII 16},
  pages={435--452},
  year={2020},
  organization={Springer}
}

@inproceedings{zhen2020joint,
  title={Joint semantic segmentation and boundary detection using iterative pyramid contexts},
  author={Zhen, Mingmin and Wang, Jinglu and Zhou, Lei and Li, Shiwei and Shen, Tianwei and Shang, Jiaxiang and Fang, Tian and Quan, Long},
  booktitle={Proceedings of the IEEE/CVF Conference on Computer Vision and Pattern Recognition},
  pages={13666--13675},
  year={2020}
}

@article{xiao2023baseg,
  title={BASeg: Boundary aware semantic segmentation for autonomous driving},
  author={Xiao, Xiaoyang and Zhao, Yuqian and Zhang, Fan and Luo, Biao and Yu, Lingli and Chen, Baifan and Yang, Chunhua},
  journal={Neural Networks},
  volume={157},
  pages={460--470},
  year={2023},
  publisher={Elsevier}
}

@article{van2017neural,
  title={Neural discrete representation learning},
  author={Van Den Oord, Aaron and Vinyals, Oriol and others},
  journal={Advances in neural information processing systems},
  volume={30},
  year={2017}
}

@inproceedings{brooks2019unprocessing,
  title={Unprocessing images for learned raw denoising},
  author={Brooks, Tim and Mildenhall, Ben and Xue, Tianfan and Chen, Jiawen and Sharlet, Dillon and Barron, Jonathan T},
  booktitle={Proceedings of the IEEE/CVF conference on computer vision and pattern recognition},
  pages={11036--11045},
  year={2019}
}

@article{binas2017ddd17,
  title={DDD17: End-to-end DAVIS driving dataset},
  author={Binas, Jonathan and Neil, Daniel and Liu, Shih-Chii and Delbruck, Tobi},
  journal={arXiv preprint arXiv:1711.01458},
  year={2017}
}

@article{gehrig2021dsec,
  title={Dsec: A stereo event camera dataset for driving scenarios},
  author={Gehrig, Mathias and Aarents, Willem and Gehrig, Daniel and Scaramuzza, Davide},
  journal={IEEE Robotics and Automation Letters},
  volume={6},
  number={3},
  pages={4947--4954},
  year={2021},
  publisher={IEEE}
}

@inproceedings{hu2021v2e,
  title={v2e: From video frames to realistic DVS events},
  author={Hu, Yuhuang and Liu, Shih-Chii and Delbruck, Tobi},
  booktitle={Proceedings of the IEEE/CVF conference on computer vision and pattern recognition},
  pages={1312--1321},
  year={2021}
}

@inproceedings{Dosovitskiy17,
  title = {{CARLA}: {An} Open Urban Driving Simulator},
  author = {Alexey Dosovitskiy and German Ros and Felipe Codevilla and Antonio Lopez and Vladlen Koltun},
  booktitle = {Proceedings of the 1st Annual Conference on Robot Learning},
  pages = {1--16},
  year = {2017}
}

@inproceedings{alonso2019ev,
  title={EV-SegNet: Semantic segmentation for event-based cameras},
  author={Alonso, Inigo and Murillo, Ana C},
  booktitle={Proceedings of the IEEE/CVF Conference on Computer Vision and Pattern Recognition Workshops},
  pages={0--0},
  year={2019}
}

@inproceedings{wang2021evdistill,
  title={Evdistill: Asynchronous events to end-task learning via bidirectional reconstruction-guided cross-modal knowledge distillation},
  author={Wang, Lin and Chae, Yujeong and Yoon, Sung-Hoon and Kim, Tae-Kyun and Yoon, Kuk-Jin},
  booktitle={Proceedings of the IEEE/CVF Conference on Computer Vision and Pattern Recognition},
  pages={608--619},
  year={2021}
}

@inproceedings{sun2022ess,
  title={Ess: Learning event-based semantic segmentation from still images},
  author={Sun, Zhaoning and Messikommer, Nico and Gehrig, Daniel and Scaramuzza, Davide},
  booktitle={European Conference on Computer Vision},
  pages={341--357},
  year={2022},
  organization={Springer}
}

@article{jia2023event,
  title={Event-based semantic segmentation with posterior attention},
  author={Jia, Zexi and You, Kaichao and He, Weihua and Tian, Yang and Feng, Yongxiang and Wang, Yaoyuan and Jia, Xu and Lou, Yihang and Zhang, Jingyi and Li, Guoqi and others},
  journal={IEEE Transactions on Image Processing},
  volume={32},
  pages={1829--1842},
  year={2023},
  publisher={IEEE}
}

@inproceedings{zhang2021issafe,
  title={ISSAFE: Improving semantic segmentation in accidents by fusing event-based data},
  author={Zhang, Jiaming and Yang, Kailun and Stiefelhagen, Rainer},
  booktitle={2021 IEEE/RSJ International Conference on Intelligent Robots and Systems (IROS)},
  pages={1132--1139},
  year={2021},
  organization={IEEE}
}

@inproceedings{biswas2024halsie,
  title={Halsie: Hybrid approach to learning segmentation by simultaneously exploiting image and event modalities},
  author={Biswas, Shristi Das and Kosta, Adarsh and Liyanagedera, Chamika and Apolinario, Marco and Roy, Kaushik},
  booktitle={2024 IEEE/CVF Winter Conference on Applications of Computer Vision (WACV)},
  pages={5952--5962},
  year={2024},
  organization={IEEE}
}

@article{xie2024eisnet,
  title={EISNet: A Multi-Modal Fusion Network for Semantic Segmentation with Events and Images},
  author={Xie, Bochen and Deng, Yongjian and Shao, Zhanpeng and Li, Youfu},
  journal={IEEE Transactions on Multimedia},
  year={2024},
  publisher={IEEE}
}

@article{zhang2023cmx,
  title={CMX: Cross-modal fusion for RGB-X semantic segmentation with transformers},
  author={Zhang, Jiaming and Liu, Huayao and Yang, Kailun and Hu, Xinxin and Liu, Ruiping and Stiefelhagen, Rainer},
  journal={IEEE Transactions on intelligent transportation systems},
  year={2023},
  publisher={IEEE}
}

@inproceedings{zhang2023delivering,
  title={Delivering arbitrary-modal semantic segmentation},
  author={Zhang, Jiaming and Liu, Ruiping and Shi, Hao and Yang, Kailun and Rei{\ss}, Simon and Peng, Kunyu and Fu, Haodong and Wang, Kaiwei and Stiefelhagen, Rainer},
  booktitle={Proceedings of the IEEE/CVF Conference on Computer Vision and Pattern Recognition},
  pages={1136--1147},
  year={2023}
}

@inproceedings{jing2024hpl,
  title={HPL-ESS: Hybrid Pseudo-Labeling for Unsupervised Event-based Semantic Segmentation},
  author={Jing, Linglin and Ding, Yiming and Gao, Yunpeng and Wang, Zhigang and Yan, Xu and Wang, Dong and Schaefer, Gerald and Fang, Hui and Zhao, Bin and Li, Xuelong},
  booktitle={Proceedings of the IEEE/CVF Conference on Computer Vision and Pattern Recognition},
  pages={23128--23137},
  year={2024}
}

@inproceedings{zhang2020learning,
  title={Learning to see in the dark with events},
  author={Zhang, Song and Zhang, Yu and Jiang, Zhe and Zou, Dongqing and Ren, Jimmy and Zhou, Bin},
  booktitle={Computer Vision--ECCV 2020: 16th European Conference, Glasgow, UK, August 23--28, 2020, Proceedings, Part XVIII 16},
  pages={666--682},
  year={2020},
  organization={Springer}
}

@inproceedings{zhou2021delieve,
  title={DeLiEve-Net: Deblurring low-light images with light streaks and local events},
  author={Zhou, Chu and Teng, Minggui and Han, Jin and Xu, Chao and Shi, Boxin},
  booktitle={Proceedings of the IEEE/CVF International Conference on Computer Vision},
  pages={1155--1164},
  year={2021}
}

@article{zhou2023deblurring,
  title={Deblurring low-light images with events},
  author={Zhou, Chu and Teng, Minggui and Han, Jin and Liang, Jinxiu and Xu, Chao and Cao, Gang and Shi, Boxin},
  journal={International Journal of Computer Vision},
  volume={131},
  number={5},
  pages={1284--1298},
  year={2023},
  publisher={Springer}
}

@inproceedings{shi2023even,
  title={EVEN: An event-based framework for monocular depth estimation at adverse night conditions},
  author={Shi, Peilun and Peng, Jiachuan and Qiu, Jianing and Ju, Xinwei and Lo, Frank Po Wen and Lo, Benny},
  booktitle={2023 IEEE International Conference on Robotics and Biomimetics (ROBIO)},
  pages={1--7},
  year={2023},
  organization={IEEE}
}

@inproceedings{liang2023coherent,
  title={Coherent event guided low-light video enhancement},
  author={Liang, Jinxiu and Yang, Yixin and Li, Boyu and Duan, Peiqi and Xu, Yong and Shi, Boxin},
  booktitle={Proceedings of the IEEE/CVF International Conference on Computer Vision},
  pages={10615--10625},
  year={2023}
}

@inproceedings{liu2023low,
  title={Low-light video enhancement with synthetic event guidance},
  author={Liu, Lin and An, Junfeng and Liu, Jianzhuang and Yuan, Shanxin and Chen, Xiangyu and Zhou, Wengang and Li, Houqiang and Wang, Yan Feng and Tian, Qi},
  booktitle={Proceedings of the AAAI Conference on Artificial Intelligence},
  pages={1692--1700},
  year={2023}
}

@article{jiang2023event,
  title={Event-based low-illumination image enhancement},
  author={Jiang, Yu and Wang, Yuehang and Li, Siqi and Zhang, Yongji and Zhao, Minghao and Gao, Yue},
  journal={IEEE Transactions on Multimedia},
  year={2023},
  publisher={IEEE}
}

@inproceedings{li2024event,
  title={Event-assisted Low-Light Video Object Segmentation},
  author={Li, Hebei and Wang, Jin and Yuan, Jiahui and Li, Yue and Weng, Wenming and Peng, Yansong and Zhang, Yueyi and Xiong, Zhiwei and Sun, Xiaoyan},
  booktitle={Proceedings of the IEEE/CVF Conference on Computer Vision and Pattern Recognition},
  pages={3250--3259},
  year={2024}
}

@inproceedings{wang2022multimodal,
  title={Multimodal token fusion for vision transformers},
  author={Wang, Yikai and Chen, Xinghao and Cao, Lele and Huang, Wenbing and Sun, Fuchun and Wang, Yunhe},
  booktitle={Proceedings of the IEEE/CVF conference on computer vision and pattern recognition},
  pages={12186--12195},
  year={2022}
}

@inproceedings{DBLP:conf/icml/JiaG0W00024,
  author       = {Ding Jia and
                  Jianyuan Guo and
                  Kai Han and
                  Han Wu and
                  Chao Zhang and
                  Chang Xu and
                  Xinghao Chen},
  title        = {GeminiFusion: Efficient Pixel-wise Multimodal Fusion for Vision Transformer},
  booktitle    = {Forty-first International Conference on Machine Learning, {ICML} 2024,
                  Vienna, Austria, July 21-27, 2024},
  publisher    = {OpenReview.net},
  year         = {2024},
  url          = {https://openreview.net/forum?id=Zsz9Pdfvtg},
  bibsource    = {dblp computer science bibliography, https://dblp.org}
}

@inproceedings{DBLP:conf/iclr/Zhou0LYDSY24,
  author       = {Hanyu Zhou and
                  Yi Chang and
                  Haoyue Liu and
                  Wending Yan and
                  Yuxing Duan and
                  Zhiwei Shi and
                  Luxin Yan},
  title        = {Exploring the Common Appearance-Boundary Adaptation for Nighttime
                  Optical Flow},
  booktitle    = {The Twelfth International Conference on Learning Representations,
                  {ICLR} 2024, Vienna, Austria, May 7-11, 2024},
  publisher    = {OpenReview.net},
  year         = {2024},
  url          = {https://openreview.net/forum?id=776lhoaulC},
  bibsource    = {dblp computer science bibliography, https://dblp.org}
}

@inproceedings{liu2022cross,
  title={Cross-Modal Discrete Representation Learning},
  author={Liu, Alex and Jin, SouYoung and Lai, Cheng-I and Rouditchenko, Andrew and Oliva, Aude and Glass, James},
  booktitle={Proceedings of the 60th Annual Meeting of the Association for Computational Linguistics (Volume 1: Long Papers)},
  pages={3013--3035},
  year={2022}
}

@article{zheng2024unicode,
  title={Unicode: Learning a unified codebook for multimodal large language models},
  author={Zheng, Sipeng and Zhou, Bohan and Feng, Yicheng and Wang, Ye and Lu, Zongqing},
  journal={arXiv preprint arXiv:2403.09072},
  year={2024}
}

@article{xia2024achieving,
  title={Achieving cross modal generalization with multimodal unified representation},
  author={Xia, Yan and Huang, Hai and Zhu, Jieming and Zhao, Zhou},
  journal={Advances in Neural Information Processing Systems},
  volume={36},
  year={2024}
}

@article{lichtsteiner2008128,
  title={A 128 $\times$ 128 120 dB 15 $\mu$s latency asynchronous temporal contrast vision sensor},
  author={Lichtsteiner, Patrick and Posch, Christoph and Delbruck, Tobi},
  journal={IEEE journal of solid-state circuits},
  volume={43},
  number={2},
  pages={566--576},
  year={2008},
  publisher={IEEE}
}

@article{brandli2014240,
  title={A 240 $\times$ 180 130 db 3 $\mu$s latency global shutter spatiotemporal vision sensor},
  author={Brandli, Christian and Berner, Raphael and Yang, Minhao and Liu, Shih-Chii and Delbruck, Tobi},
  journal={IEEE Journal of Solid-State Circuits},
  volume={49},
  number={10},
  pages={2333--2341},
  year={2014},
  publisher={IEEE}
}

@article{taverni2018front,
  title={Front and back illuminated dynamic and active pixel vision sensors comparison},
  author={Taverni, Gemma and Moeys, Diederik Paul and Li, Chenghan and Cavaco, Celso and Motsnyi, Vasyl and Bello, David San Segundo and Delbruck, Tobi},
  journal={IEEE Transactions on Circuits and Systems II: Express Briefs},
  volume={65},
  number={5},
  pages={677--681},
  year={2018},
  publisher={IEEE}
}

@article{rebecq2019high,
  title={High speed and high dynamic range video with an event camera},
  author={Rebecq, Henri and Ranftl, Ren{\'e} and Koltun, Vladlen and Scaramuzza, Davide},
  journal={IEEE transactions on pattern analysis and machine intelligence},
  volume={43},
  number={6},
  pages={1964--1980},
  year={2019},
  publisher={IEEE}
}

@inproceedings{stoffregen2020reducing,
  title={Reducing the sim-to-real gap for event cameras},
  author={Stoffregen, Timo and Scheerlinck, Cedric and Scaramuzza, Davide and Drummond, Tom and Barnes, Nick and Kleeman, Lindsay and Mahony, Robert},
  booktitle={Computer Vision--ECCV 2020: 16th European Conference, Glasgow, UK, August 23--28, 2020, Proceedings, Part XXVII 16},
  pages={534--549},
  year={2020},
  organization={Springer}
}

@inproceedings{pan2019bringing,
  title={Bringing a blurry frame alive at high frame-rate with an event camera},
  author={Pan, Liyuan and Scheerlinck, Cedric and Yu, Xin and Hartley, Richard and Liu, Miaomiao and Dai, Yuchao},
  booktitle={Proceedings of the IEEE/CVF Conference on Computer Vision and Pattern Recognition},
  pages={6820--6829},
  year={2019}
}

@inproceedings{jiang2020learning,
  title={Learning event-based motion deblurring},
  author={Jiang, Zhe and Zhang, Yu and Zou, Dongqing and Ren, Jimmy and Lv, Jiancheng and Liu, Yebin},
  booktitle={Proceedings of the IEEE/CVF Conference on Computer Vision and Pattern Recognition},
  pages={3320--3329},
  year={2020}
}

@inproceedings{shang2021bringing,
  title={Bringing events into video deblurring with non-consecutively blurry frames},
  author={Shang, Wei and Ren, Dongwei and Zou, Dongqing and Ren, Jimmy S and Luo, Ping and Zuo, Wangmeng},
  booktitle={Proceedings of the IEEE/CVF International Conference on Computer Vision},
  pages={4531--4540},
  year={2021}
}

@article{xie2021segformer,
  title={SegFormer: Simple and efficient design for semantic segmentation with transformers},
  author={Xie, Enze and Wang, Wenhai and Yu, Zhiding and Anandkumar, Anima and Alvarez, Jose M and Luo, Ping},
  journal={Advances in neural information processing systems},
  volume={34},
  pages={12077--12090},
  year={2021}
}

@inproceedings{DBLP:conf/iclr/LoshchilovH19,
  author       = {Ilya Loshchilov and
                  Frank Hutter},
  title        = {Decoupled Weight Decay Regularization},
  booktitle    = {7th International Conference on Learning Representations, {ICLR} 2019,
                  New Orleans, LA, USA, May 6-9, 2019},
  publisher    = {OpenReview.net},
  year         = {2019},
  url          = {https://openreview.net/forum?id=Bkg6RiCqY7},
  bibsource    = {dblp computer science bibliography, https://dblp.org}
}

@inproceedings{smith2017cyclical,
  title={Cyclical learning rates for training neural networks},
  author={Smith, Leslie N},
  booktitle={2017 IEEE winter conference on applications of computer vision (WACV)},
  pages={464--472},
  year={2017},
  organization={IEEE}
}

@inproceedings{zihao2018unsupervised,
  title={Unsupervised event-based optical flow using motion compensation},
  author={Zihao Zhu, Alex and Yuan, Liangzhe and Chaney, Kenneth and Daniilidis, Kostas},
  booktitle={Proceedings of the European Conference on Computer Vision (ECCV) Workshops},
  pages={711--714},
  year={2018},
}

@inproceedings{ning2023all,
  title={All in tokens: Unifying output space of visual tasks via soft token},
  author={Ning, Jia and Li, Chen and Zhang, Zheng and Wang, Chunyu and Geng, Zigang and Dai, Qi and He, Kun and Hu, Han},
  booktitle={Proceedings of the IEEE/CVF International Conference on Computer Vision},
  pages={19900--19910},
  year={2023}
}

@inproceedings{chen2023generative,
  title={Generative semantic segmentation},
  author={Chen, Jiaqi and Lu, Jiachen and Zhu, Xiatian and Zhang, Li},
  booktitle={Proceedings of the IEEE/CVF Conference on Computer Vision and Pattern Recognition},
  pages={7111--7120},
  year={2023}
}

@article{gallego2020event,
  title={Event-based vision: A survey},
  author={Gallego, Guillermo and Delbr{\"u}ck, Tobi and Orchard, Garrick and Bartolozzi, Chiara and Taba, Brian and Censi, Andrea and Leutenegger, Stefan and Davison, Andrew J and Conradt, J{\"o}rg and Daniilidis, Kostas and others},
  journal={IEEE transactions on pattern analysis and machine intelligence},
  volume={44},
  number={1},
  pages={154--180},
  year={2020},
  publisher={IEEE}
}

@article{bengio2013estimating,
  title={Estimating or propagating gradients through stochastic neurons for conditional computation},
  author={Bengio, Yoshua and L{\'e}onard, Nicholas and Courville, Aaron},
  journal={arXiv preprint arXiv:1308.3432},
  year={2013}
}

@article{bayer1976color,
  title={Color imaging array},
  author={Bayer, Bryce},
  journal={United States Patent, no. 3971065},
  year={1976}
}

@inproceedings{DBLP:conf/iccv/XiaZZWST23,
  author       = {Ruihao Xia and
                  Chaoqiang Zhao and
                  Meng Zheng and
                  Ziyan Wu and
                  Qiyu Sun and
                  Yang Tang},
  title        = {{CMDA:} Cross-Modality Domain Adaptation for Nighttime Semantic Segmentation},
  booktitle    = {{IEEE/CVF} International Conference on Computer Vision, {ICCV} 2023,
                  Paris, France, October 1-6, 2023},
  pages        = {21515--21524},
  publisher    = {{IEEE}},
  year         = {2023},
  url          = {https://doi.org/10.1109/ICCV51070.2023.01972},
  doi          = {10.1109/ICCV51070.2023.01972},
  bibsource    = {dblp computer science bibliography, https://dblp.org}
}

@inproceedings{qi2023e2nerf,
  title={E2nerf: Event enhanced neural radiance fields from blurry images},
  author={Qi, Yunshan and Zhu, Lin and Zhang, Yu and Li, Jia},
  booktitle={Proceedings of the IEEE/CVF International Conference on Computer Vision},
  pages={13254--13264},
  year={2023}
}

@inproceedings{geng2024event,
  title={Event-based Visible and Infrared Fusion via Multi-task Collaboration},
  author={Geng, Mengyue and Zhu, Lin and Wang, Lizhi and Zhang, Wei and Xiong, Ruiqin and Tian, Yonghong},
  booktitle={Proceedings of the IEEE/CVF Conference on Computer Vision and Pattern Recognition},
  pages={26929--26939},
  year={2024}
}

@inproceedings{zhao2024edge,
  title={Edge-Guided Fusion and Motion Augmentation for Event-Image Stereo},
  author={Zhao, Fengan and Zhou, Qianang and Xiong, Junlin},
  booktitle={European Conference on Computer Vision},
  pages={190--205},
  year={2024},
  organization={Springer}
}

@article{tao2020hierarchical,
  title={Hierarchical multi-scale attention for semantic segmentation},
  author={Tao, Andrew and Sapra, Karan and Catanzaro, Bryan},
  journal={arXiv preprint arXiv:2005.10821},
  year={2020}
}

@inproceedings{yang2023event,
  title={Event camera data pre-training},
  author={Yang, Yan and Pan, Liyuan and Liu, Liu},
  booktitle={Proceedings of the IEEE/CVF international conference on computer vision},
  pages={10699--10709},
  year={2023}
}

@inproceedings{yao2024sam,
  title={Sam-event-adapter: Adapting segment anything model for event-rgb semantic segmentation},
  author={Yao, Bowen and Deng, Yongjian and Liu, Yuhan and Chen, Hao and Li, Youfu and Yang, Zhen},
  booktitle={2024 IEEE International Conference on Robotics and Automation (ICRA)},
  pages={9093--9100},
  year={2024},
  organization={IEEE}
}

@article{wu2025cm3ae,
  title={CM3AE: A Unified RGB Frame and Event-Voxel/-Frame Pre-training Framework},
  author={Wu, Wentao and Wang, Xiao and Li, Chenglong and Jiang, Bo and Tang, Jin and Luo, Bin and Liu, Qi},
  journal={arXiv preprint arXiv:2504.12576},
  year={2025}
}

@inproceedings{kim2024missing,
  title={Missing modality prediction for unpaired multimodal learning via joint embedding of unimodal models},
  author={Kim, Donggeun and Kim, Taesup},
  booktitle={European Conference on Computer Vision},
  pages={171--187},
  year={2024},
  organization={Springer}
}

@inproceedings{lang2025retrieval,
  title={Retrieval-augmented dynamic prompt tuning for incomplete multimodal learning},
  author={Lang, Jian and Cheng, Zhangtao and Zhong, Ting and Zhou, Fan},
  booktitle={Proceedings of the AAAI Conference on Artificial Intelligence},
  pages={18035--18043},
  year={2025}
}

@inproceedings{zhao2025eseg,
  title={ESEG: Event-Based Segmentation Boosted by Explicit Edge-Semantic Guidance},
  author={Zhao, Yucheng and Lyu, Gengyu and Li, Ke and Wang, Zihao and Chen, Hao and Yang, Zhen and Deng, Yongjian},
  booktitle={Proceedings of the AAAI Conference on Artificial Intelligence},
  pages={10510--10518},
  year={2025}
}
}

\clearpage

\appendix

\section*{Appendix}

\label{sec:supp}

\section{Details of datasets}

\label{sec:appendixA}

Three extreme-condition datasets are constructed and used for testing in our work, namely DERS-XS, DERS-XR, and DSEC-Xtrm. Among them, DERS-XS and DSEC-Xtrm are synthetic datasets, and DERS-XR is a real-world dataset. In addition, a common dataset of normal conditions, namely DSEC-Semantic is used for testing. By leveraging edge-awareness, our method can effectively obtain the common features of heterogeneous event and RGB under unified semantic space and jointly optimize them. Results show that our method outperforms existing event-RGB segmentation methods and possesses superior resilience in the case of modality imbalance and failure under extreme conditions.

\begin{table*}[h]
\centering
\captionsetup{font=small}
\caption{Comparison between datasets.}\label{tab:comp_datasets}
\setlength\tabcolsep{5pt}
\renewcommand{\arraystretch}{1}
\resizebox{1\textwidth}{!}{
\begin{tabular}{lcccccccc}
 \toprule
 Datasets & \#Train & \#Validation & \#Test &  Real/Syn. & True/Pseudo-lbl. & Fine/Coarse-lbl. & Avg. Pixel Val. & Avg. \#Events \\
 \midrule
 DERS-XS & 2016 & 144 & 1080 & Synthetic & True-label & Fine-label & 6.16 & 70490.88 \\
 DERS-XR & 120 & N/A & 120 & Real-world & True-label & Coarse-label & 20.58 & 12784.76 \\
 DSEC-Semantic & 8082 & N/A & 2809 & Real-world & Pseudo-label & Fine-label & 75.13 & 608162.26 \\
 DSEC-Xtrm & 8082 & N/A & 2809 & Synthetic & Pseudo-label & Fine-label & 4.75 & 689098.25 \\
 \bottomrule
\end{tabular}
}
\end{table*}

\subsection{DERS-XS}

Our synthetic dataset DERS-XS is constructed based on the CARLA simulator \cite{Dosovitskiy17} and v2e simulator \cite{hu2021v2e}, containing 270 frame sequences. We first obtain canonical RGB frames with labels from CARLA simulator. For CARLA simulation process, we first load six pre-made maps, namely \textit{Town01, Town02, Town03, Town04, Town05, and Town10}. Then we apply fifteen different types of weather conditions on each map. For each weather and map combination, we record two sequences of 1200 frames at a frame rate of 20 fps. The spatial size of each frame is 640 \texttimes~360.

The CARLA simulated segmentation labels contain 23 categories, and the specific category names are \textit{unlabeled, building, fence, other, pedestrian, pole, roadline, road, sidewalk, vegetation, vehicles, wall, traffic sign, sky, ground, bridge, rail track, guard rail, traffic light, static, dynamic, water, and terrain}. For experiments, we merge and transform the above 23 categories into 11 categories, in order to match the categories setting of DSEC-Semantic \cite{gehrig2021dsec, sun2022ess}. The 11 categories are \textit{background, building, fence, person, pole, road, sidewalk, vegetation, car, wall, and traffic sign}. Categories that do not exist in DSEC-Semantic are set to 255 and ignored during the training and testing process.

We then obtain noisy events from v2e simulator based on the CARLA simulated frames. For v2e simulation process, the positive threshold and the negative threshold are both set as 0.2. The 1-std deviation threshold variation is set as 0.05. The cutoff frequency is set as 30, and the leak event rate per pixel is set as 0.1. The shot noise rate is set as 5.0. The refractory period is set as 0.0005. 

\begin{figure*}[t]
    \centering
    \includegraphics[width=1.0\linewidth]{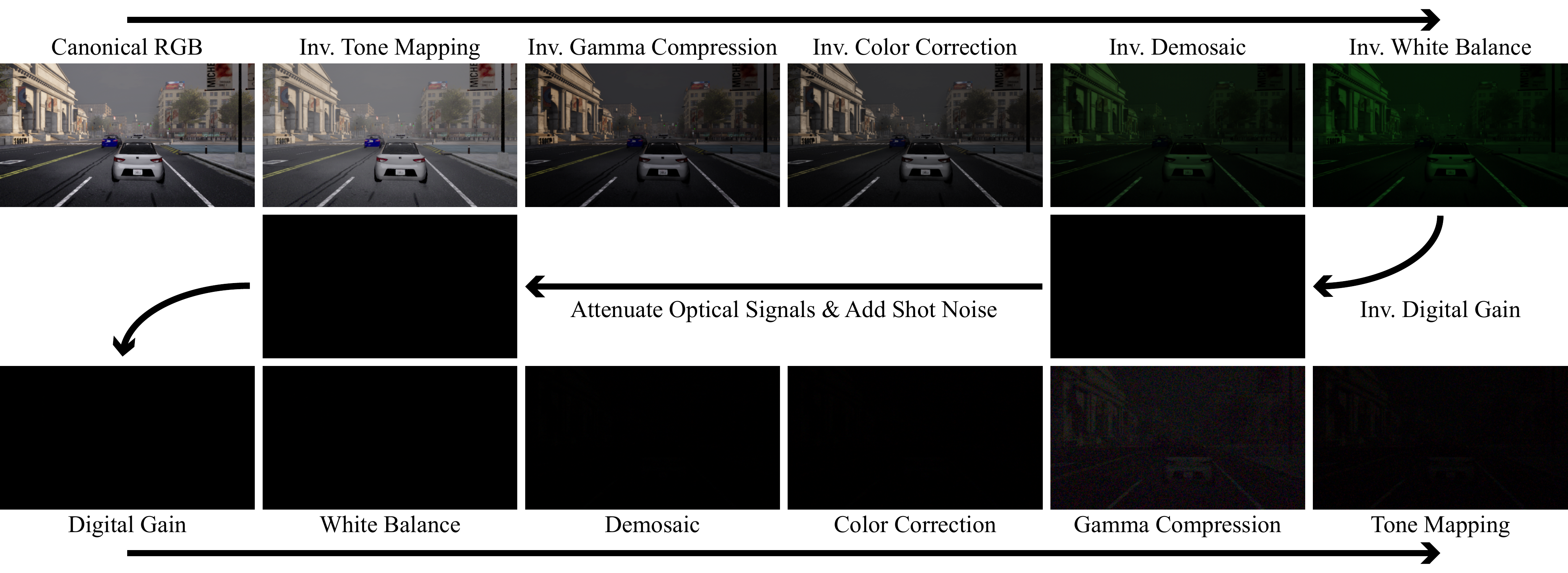}
    \captionsetup{font=small}
    \caption{\textbf{Obtaining low-light RGB images using ISP with its inversion process.} We implement the ISP with its inversion process, including digital gain, white balance, demosaic, color correction, gamma compression, and tone mapping. Optical Signals are attenuated, and shot noise is added on RAW domain images. The process is used to simulate low-light RGB images in DERS-XS and DSEC-Xtrm.}
    \label{fig:SUPP-ISP}
\end{figure*}

We implement the Image Signal Processing (ISP) pipeline and its inverse process, referred to as the ISP unprocessing technique in \cite{brooks2019unprocessing}, which includes digital gain, white balance, demosaicing, color correction, gamma compression, and tone mapping.
By utilizing the inversion of ISP, we first convert CARLA-simulated canonical RGB frames into Bayer-pattern BGGR RAW images \cite{bayer1976color}. As shown in \cref{fig:SUPP-ISP}, we attenuate the optical signals and add shot noise on RAW images, and then process RAW images by ISP to obtain low-light images. 

We sample the 1200-frame sequences at the intervals of 100 frames, taking 12 frames from each sequence while discarding the rest. We divide the 270 sequences into three parts, of which the training set has 168 sequences, the validation set has 12 sequences, and the test set has 90 sequences. For training process of DERS-XS, we use the training set for training, and the validation set for saving the model with the best validation mIoU. The test set is used only during the testing process.

\subsection{DERS-XR}

Our real-world dataset DERS-XR is captured by a DAVIS-346 \cite{taverni2018front}, with the spatial size of 346 \texttimes~260. We use the camera to capture the events with APS frames of 20 fps simultaneously under extreme lighting conditions. We capture a total of 27 frame sequences, sampling them at intervals of 100 frames, and finally obtain 240 frames of images with very different contents. We manually annotate semantic labels for the sampled 240 frames, 
and the annotated categories are the same as the transformed 11 categories of DSEC-XS. We randomly select 120 frames for fine-tuning and the remaining 120 frames for testing. For experiments conducted on DERS-XR, we save the last epoch fine-tuning model for testing.

\subsection{DSEC-Xtrm}

\begin{wrapfigure}{r}{0.5\textwidth}
    \centering
    \includegraphics[width=1.0\linewidth]{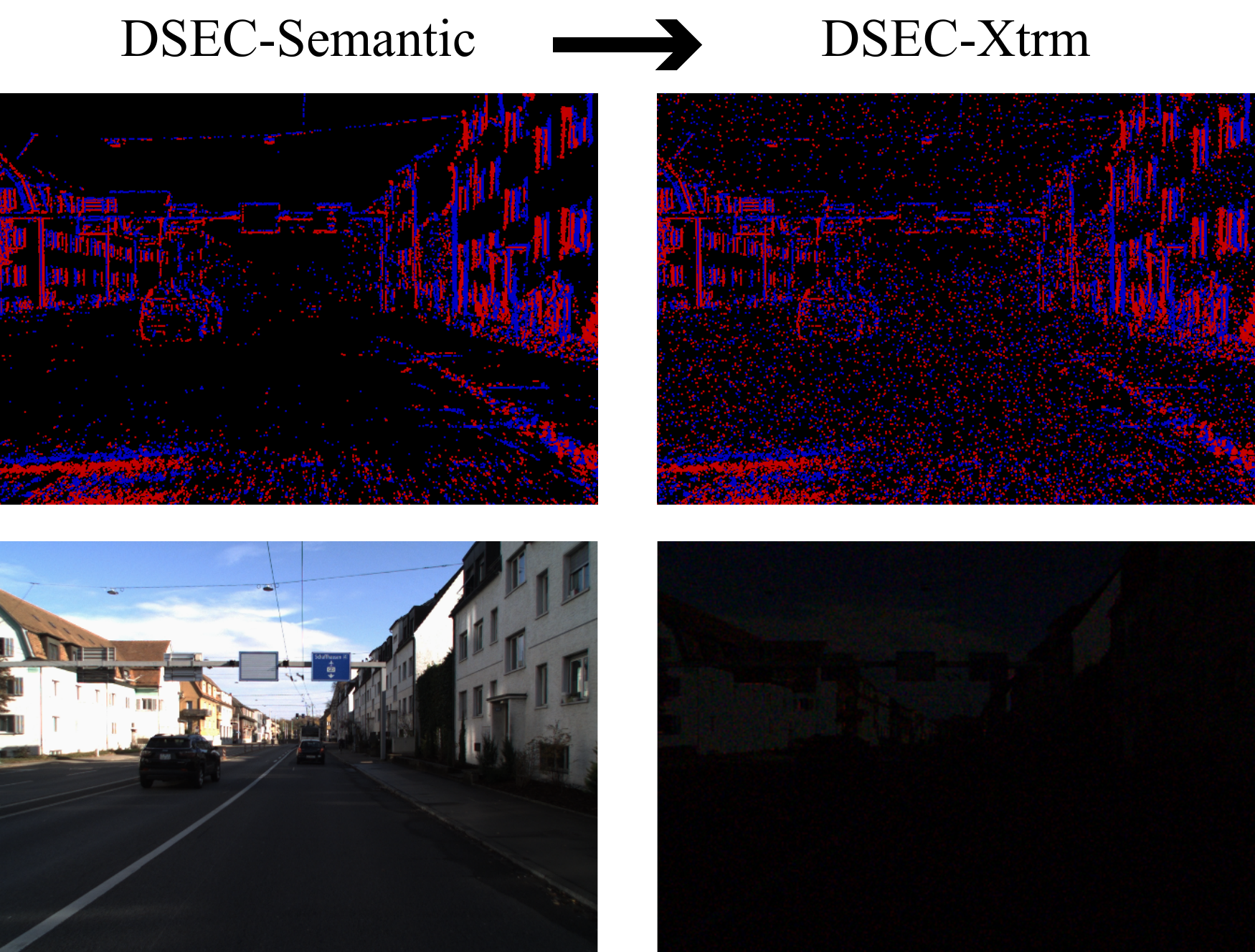}
    \captionsetup{font=small}
    \caption{DSEC-Semantic and DSEC-Xtrm.}
    \label{fig:SUPP-2}
\end{wrapfigure}

DSEC-Xtrm is simulated and converted from the real-world dataset DSEC-Semantic.  A sample of DSEC-Semantic and DSEC-Xtrm is shown in \cref{fig:SUPP-2}. DSEC-Semantic includes 11 sequences with pseudo-labels of 19 categories and 11 categories. We generate pure noise events by modifying the source codes of v2e simulator and overlaying them to the event sequence of DSEC-Semantic to obtain noisy events in DSEC-Xtrm. The shot noise rate is set as 10.0 Hz. To obtain low-light images, we apply the same ISP and ISP-inversion process, unprocessing RGB images to RAW domain, attenuating the optical signals and adding shot noise on the unprocessed RAW images, and then process the RAW images to RGB low-light images by ISP. We sample the 11 sequences at the intervals of 2 frames and discard the first 6 frames of
each sequence, following the same setting with \cite{sun2022ess}. The simulated low-light images and noisy events combined with the original 11-categories labels of DSEC-Semantic together constitute the final DSEC-Xtrm dataset.

\subsection{Discussion on datasets}

This paper uses the DSEC-Semantic and constructs three datasets, each of which has its own properties. \Cref{tab:comp_datasets} compares the properties of different datasets, including the number of image-events-pair in the training set, validationset and test set, whether the data is real-world or synthetic, whether the label is true or pseudo, whether the label is fine or coarse, and the average pixel value of images and average number of events of each dataset. Based on the different properties of different datasets, we can use them in different experimental settings to test the model comprehensively.

The DERS-XS only contains synthetic data, but \textbf{DERS-XS has the largest amount of data with the true fine-grained label, thus DERS-XS can be a standard benchmark for comparative experiments}. The DERS-XR is a real-world dataset with the most realistic data distribution, however, it is difficult to annotate it accurately, and for this reason, we only coarsely annotate a small amount of data of DERS-XR. \textbf{Since it is difficult to annotate DERS-XR, only a small amount of data is annotated, thus we only use DERS-XR for fine-tuning and testing.} 

The DSEC-Semantic is a real-world dataset under normal conditions, however, the labels are pseudo labels based on RGB images only. Although it is used for testing in many works, \textbf{using DSEC-Semantic as a benchmark for multi-modality semantic segmentation is defective}. Thus, we simulate the extreme version DSEC-Xtrm from DSEC-Semantic. For DSEC-Xtrm, the pseudo labels from DSEC-Semantic are no longer a defect for being a benchmark for multi-modality semantic segmentation. \textbf{Therefore, when testing on the DSEC-Semantic and DSEC-Xtrm, we can compare the performance degradation on the two datasets for different methods.} The performance degradation on DSEC-Semantic and DSEC-Xtrm can illustrate the robustness and resilience of different methods.

\section{Details of training settings}

\label{sec:appendixB}

For all datasets, we use AdamW \cite{DBLP:conf/iclr/LoshchilovH19} optimizer with a weight decay of 0.01, and the learning rate of decoder is 10 times the basic learning rate.
For DERS-XS, we train our model on two NVIDIA RTX 3090 GPUs for 300 epochs, and the batch size is 16 on each GPU. The basic learning rate (LR) is 6 \texttimes~10$^{-\text{5}}$, which is scheduled by a CyclicLR \cite{smith2017cyclical} scheduler with a maximum learning rate 1.6 \texttimes~LR and a triangular cycle of 10 epochs. 
For DERS-XR, we fine-tune our model for 50 epochs on a single GPU with a batch size of 2, based on the best model trained on DERS-XS. The basic learning rate is 6 \texttimes~10$^{-\text{5}}$, which is scheduled by a WarmupPolyLR scheduler with power of 0.9 and warmup epochs of 10. 
For DSEC-Semantic and DSEC-Xtrm, we follow settings from DERS-XS with epochs of 60.
We input events with a 50 ms interval for DERS-XS and DERS-XR, and a count of 50000 events for DSEC-Semantic and DSEC-Xtrm. 

\section{Details of event-edge statistics}
\label{sec:appendixC}

\begin{figure}[t]
  \centering
  \begin{minipage}[t]{0.5\linewidth}
    \centering
    \includegraphics[width=\linewidth]{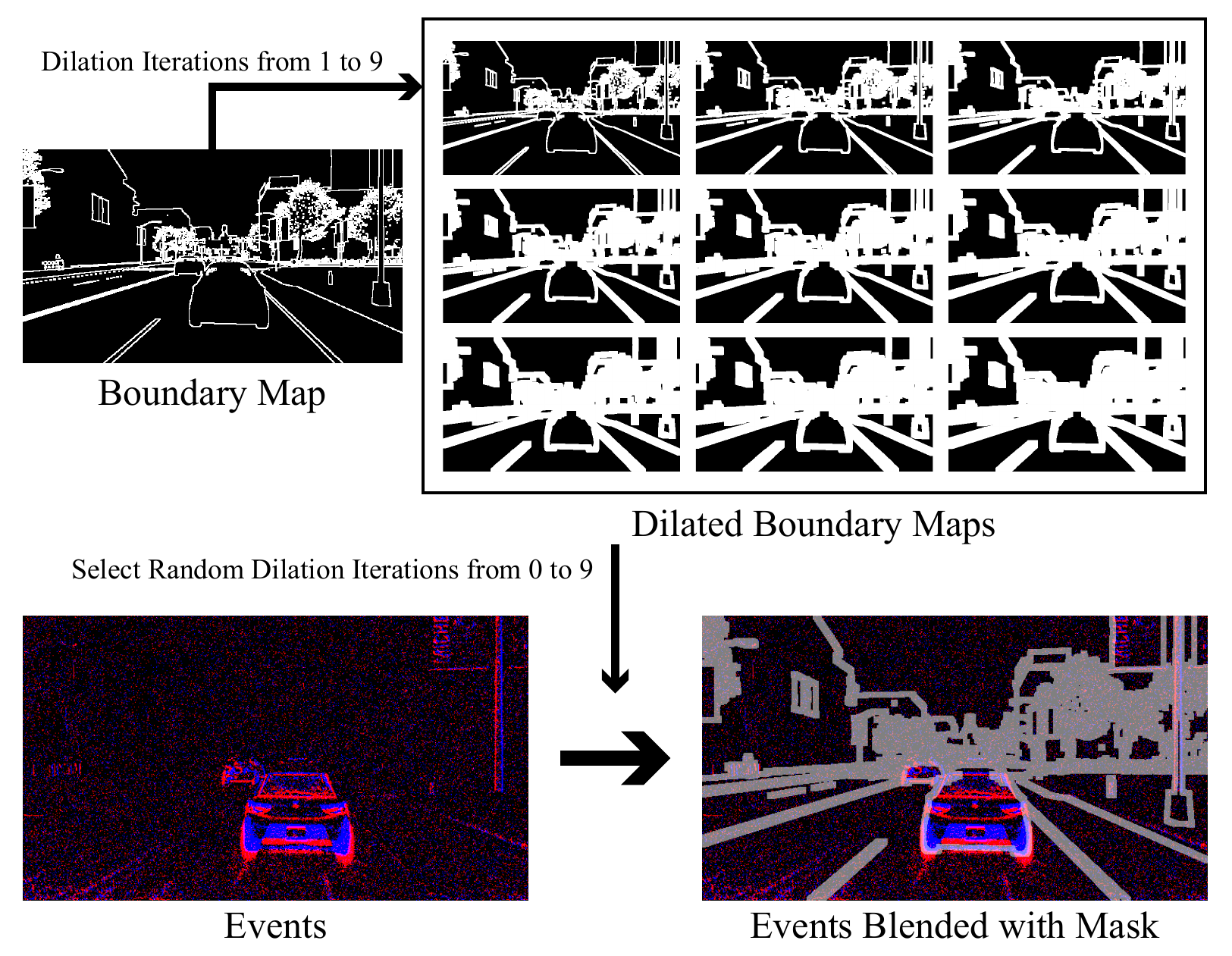}
    \captionsetup{font=small}
    \caption{Details of boundary map dilation.}
    \label{fig:SUPP-Edge-Relation}
  \end{minipage}%
  \hfill
  \begin{minipage}[t]{0.5\linewidth}
    \centering
    \includegraphics[width=\linewidth]{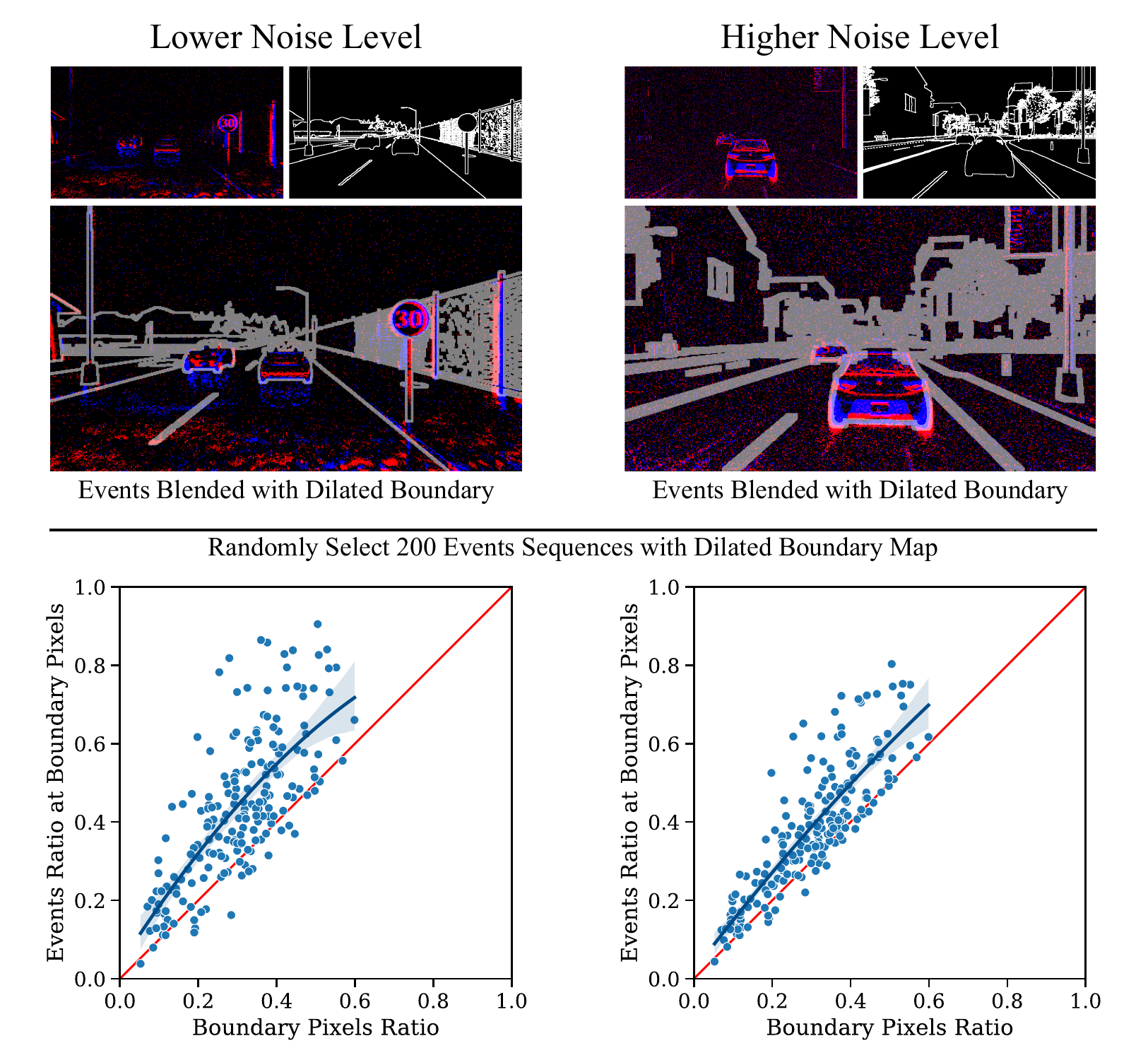}
    \captionsetup{font=small}
    \caption{Event-edge statistics under two noise levels.}
    \label{fig:SUPP-Edge-Relation-2}
  \end{minipage}
\end{figure}

For the statistical process, we randomly select 200 event sequences of 50 ms accompanied by their boundary maps from DERS-XS as sample. We aim to count the ratio of edge pixels to all pixels of the whole plane, and the ratio of events falling on edge pixels to all events of the whole sequence. As shown in \cref{fig:SUPP-Edge-Relation}, in order to make the ratio of edge pixels distributed in a larger range, we first dilate the boundary map with a 3 \texttimes~3 kernel with a random number of iterations in the range of 10. 
% \cite{opencv_library}. 
We draw a scatter plot to show the correlation between the two ratios.

As the boundary map dilates, the ratio of edge pixels increases, and the ratio of events falling on edge pixels also increases synchronously. Considering the extreme cases, when the ratio of edge pixels is 0, there is no events falling on edge pixels, then the ratio of events is also 0; when the boundary map dilates to the whole image, the ratio of edge pixels is 1, then all events fall on edge pixels, the ratio of events is 1. However, for the cases where the ratio of edge pixels is between 0 and 1, for most samples, the ratio of events falling on edge pixels is greater than the ratio of edge pixels. This proves the fact that events tend to cluster at the areas of semantic edge. This phenomenon exhibits a strong correlation between events and semantic edge.

We also count the two ratios under two noise levels of events. We use v2e simulator to simulate a version of event sequences that are less noisy than the event sequences of DERS-XS. We simultaneously count the event ratios corresponding to the two noise versions of the event sequences, and draw scatter plots for the lower noise level event sequences and the higher noise level event sequences (\ie DERS-XS). We compare the scatter plots of the two noise levels. 
As shown in \cref{fig:SUPP-Edge-Relation-2}, as the noise level increases, the ratio of edge pixels to all pixels and the ratio of events falling on edge pixels to all events tend to be equal. The correlation curve for the higher noise level case is still a concave curve, thus the correlation is still maintained even under the influence of high noise. This ensures the resilience of our method in the case of modality imbalance or failure under extreme conditions.

\section{Details of edge dictionary training process}
\label{sec:appendixD}

The training of our edge dictionary is a separate stage from the training of the segmentation model. Through this separate training stage, we obtain the pre-trained weights of edge dictionary, which represents the information of the semantic edge. We also obtain the pre-trained weights of the tokenizer, which can be used to embed the semantic edge ground truth into the discrete latent space defined by edge dictionary. The detokenizer is only used in the edge dictionary training stage, and is deprecated in the segmentation model training stage.

The tokenizer $f_T$ consists of two convolutional layers followed by ReLU, two residual blocks, and a final convolutional layer. The first two convolutional layers downsampled the inputs each with a kernel of 4 \texttimes~4 size, a stride of \textlangle 2, 2\textrangle, and a padding of \textlangle 1, 1\textrangle. The two residual blocks keep the spatial scale unchanged, each consists three convolutional layers followed by ReLU with a kernel of 3 \texttimes~3 size, a stride of \textlangle 1, 1\textrangle, and a padding of \textlangle 1, 1\textrangle. The final convolutional layers adjust the number of channels to $n$ by a convolutional layer with a kernel of 1 \texttimes~1 size. For semantic edge  $\mathbf{B} \in {\{0, 1\}}^{H \times W}$, the final produced edge embeddings $\mathbf{\Gamma} = f_T(\mathbf{B})\in \mathbb{R}^{H' \times W' \times n}$ have the downsampled spatial size $H' \times W'$, where $H' = \lfloor\frac{H}{4}\rfloor , W' = \lfloor\frac{W}{4}\rfloor$.

The detokenizer $f_{T'}$ is constructed by changing the downsampled convolutional layers to the transposed convolutional layer for upsampling and inverting all the layers in tokenizer. Thus, the detokenizer consists of a convolutional layer followed by ReLU, two residual blocks and two transposed convolutional layers followed by ReLU, and a final convolutional layer to predict the semantic edge. The kernel size, stride, and padding settings of convolutional layers and transposed convolutional layers are the same as the corresponding layers in tokenizer. The detokenizer takes the quantised embeddings $\mathbf{\Gamma}' \in \mathbb{R}^{H'\times W' \times n}$ as input, predicts the reconstructed semantic edge $\mathbf{B}' = f_{T'}(\mathbf{\Gamma}')$.

We adopt the training objective with reconstruction loss, embedding loss, and commitment loss of VQ-VAE \cite{van2017neural} as $L_{dict} = \|\mathbf{B}-\mathbf{B}'\|_2^2~+ \|v(\hat{\mathbf{K}}) - \mbox{sg}(\mathbf{\Gamma})\|_2^2+\alpha\|\mbox{sg}(v(\hat{\mathbf{K}})) - \mathbf{\Gamma}\|_2^2,$ where sg means stop gradient, and $\alpha$ is a constant of commitment loss weight, which is 0.25 in our work. To make the reconstruction loss propagate back to the tokenizer, a gradient straight-through technique \cite{bengio2013estimating} is adopted, which directly assigns the gradient from $\mathbf{\Gamma}'$ to $\mathbf{\Gamma}$. We train the edge dictionary with tokenizer and detokenizer for 3000 epochs, based on the data of DERS-XS and DSEC-Semantic. No additional information is introduced, and the pre-trained weights of tokenizer are introduced into the segmentation training stage only for the supervision in latent space, and not utilized for testing stage.

\section{Additional ablation studies}

\subsection{Edge Dictionary Domain Transferability}

Theoretically, the discrete edge dictionary learned by VQ-VAE is expected to have good transferability across datasets. As an intermediate representation, semantic edge exhibits relatively simple and consistent structures, and the latent distributions of semantic edge derived from segmentation labels tend to vary only slightly across different datasets. Therefore, we expect the performance degradation under dictionary transfer settings to be minimal.

We conduct cross-domain transferability evaluations by (i) using an edge dictionary pretrained on DSEC to evaluate on DERS-XS, and (ii) using a dictionary pretrained on DERS-XS to evaluate on DSEC-Semantic and DSEC-Xtrm. As shown in \cref{tab:sup-g1}, the performance drops are small in all cases. Specifically, the mIoU on DERS-XS drops by 0.66\%, on DSEC-Semantic by 0.10\%, and on DSEC-Xtrm by 0.21\%, compared to the original non-exchanged dictionary settings. These results suggest that the learned edge dictionary generalizes well across domains, and our method remains robust under moderate domain shifts.

\begin{table*}[t]
\centering
\captionsetup{font=small}
\caption{Ablation study on edge dictionary domain transferability.}\label{tab:sup-g1}
\setlength\tabcolsep{12pt}
\renewcommand{\arraystretch}{1}
\resizebox{0.95\linewidth}{!}{
\begin{tabular}{l c c c}
\mytoprule
  Settings & gACC(\%)\textuparrow & mACC(\%)\textuparrow & mIoU(\%)\textuparrow \\ 
 \mydmidrule
 ESC on DERS-XS (w/ edge dictionary of DSEC) & 93.23 & 74.96 & 66.44 \\
 ESC on DERS-XS & 93.27 & 75.26 & 67.10 \\ 
 \mymidrule
 ESC on DSEC-Semantic (w/ edge dictionary of DERS-XS) & 94.91 & 78.04 & 70.93 \\
 ESC on DSEC-Semantic & 94.85 & 78.61 & 71.04 \\ 
 \mymidrule
 ESC on DSEC-Xtrm (w/ edge dictionary of DERS-XS) & 88.57 & 58.00 & 50.65 \\
 ESC on DSEC-Xtrm & 88.18 & 59.45 & 50.87 \\
 \mybottomrule
\end{tabular}
}
\end{table*}

\subsection{Event Sampling Strategy}

The DSEC-Semantic event input is built by sampling a fixed number of events per voxel grid rather than a fixed time window, which follows the setting of ESS \cite{sun2022ess}. In ESS, the event input is built with 100,000 events per voxel grid. We found that a 50-ms fixed time window or 100,000 fixed number of events is relatively large, which reduces the performance of data preprocessing, and may lead to insufficient edge characteristic representation. After the above trade-offs, we decided to use 50,000 events per voxel grid for DSEC-Semantic as our event sampling strategy in our work.

To further demonstrate the impact of different event sampling strategies, we conduct experiments on DSEC-Semantic with a fixed time window of 50 ms, a fixed number of events of 100,000 per voxel grid, compared with 50,000 events per voxel grid in the main paper. As shown in \cref{tab:sup-g2}, different event sampling strategies present comparable results, with mIoU slightly lower (0.20\% and 0.19\% respectively) than the fixed number of events of 50,000 in the main paper.

\begin{table*}[t]
\centering
\captionsetup{font=small}
\caption{Ablation study on different event sampling strategies.}\label{tab:sup-g2}
\setlength\tabcolsep{20pt}
\renewcommand{\arraystretch}{1}
\resizebox{0.95\linewidth}{!}{
\begin{tabular}{l c c c}
\mytoprule
 Settings & gACC(\%)\textuparrow & mACC(\%)\textuparrow & mIoU(\%)\textuparrow \\ 
 \mymidrule
 ESC on DSEC-Semantic (50 ms) & 94.76 & 78.44 & 70.83 \\
 ESC on DSEC-Semantic (100,000 events) & 94.87 & 78.00 & 70.84 \\ 
 ESC on DSEC-Semantic (50,000 events) & 94.85 & 78.61 & 71.04 \\
 \mybottomrule
\end{tabular}
}
\end{table*}

\subsection{Deployment Efficiency and FLOPs Ablation}

We conduct additional comparative experiments on DERS-Xtrm using smaller backbones (2\texttimes~MiT-B0), reducing the FLOPs of ESC to be even lower than CMNeXt. In addition, we measure the end-to-end inference latency of CMNeXt and our ESC (both standard ESC and reduced variant) on a single NVIDIA GeForce RTX 3090 GPU with a batch size of 1. All latency measurements are performed with a fixed input size of 512\texttimes~512, with each measurement calculating the average execution time over 100 inferences, and we repeat 3 times for stability.

As shown in \cref{tab:sup-g3}, the reduced ESC variant still outperforms CMNeXt, achieving 49.06\% mIoU vs. 45.16\%, despite lower FLOPs (60.658G vs. 62.805G) and significantly fewer parameters (14.184M vs. 58.687M). This suggests that the performance gains stem from architectural design rather than merely an increased computational cost. Furthermore, as shown in \cref{tab:sup-g3}, the reduced ESC variant has an average inference latency of 29.37 ms, which is shorter than that of CMNeXt (29.79 ms), demonstrating its potential for more efficient deployment.

Above results indicate that even with lightweight backbones, our model maintains strong performance with higher inference speed, highlighting the effectiveness of our design beyond raw FLOPs. This reflects a favorable trade-off between efficiency and performance, which is essential for practical deployment in real-world systems.

\begin{table*}[t]
\centering
\captionsetup{font=small}
\caption{Ablation study on lighter backbones on DSEC-Xtrm.}\label{tab:sup-g3}
\setlength\tabcolsep{8pt}
\renewcommand{\arraystretch}{1}
\resizebox{1\linewidth}{!}{
\begin{tabular}{lccccc|cccc}
\mytoprule
    \multirow{2}{*}{Settings} & \multirow{2}{*}{gACC(\%)\textuparrow} & \multirow{2}{*}{mACC(\%)\textuparrow} & \multirow{2}{*}{mIoU(\%)\textuparrow} & \multirow{2}{*}{\#Params(M)} & \multirow{2}{*}{FLOPs(G)} & \multicolumn{4}{c}{Latency (ms)}\\
    &&&&&& \#1 & \#2 & \#3 & Avg. \\ 
    \mymidrule
    CMNeXt \cite{zhang2023delivering} & 87.04 & 52.12 & 45.16 & 58.687 & 62.805 & 29.75 & 29.78 & 29.83 & 29.79 \\
    ESC (Reduced) & 88.03 & 56.31 & 49.06 & 14.184 & 60.658 & 29.30 & 29.38 & 29.43 & 29.37 \\
    ESC (Standard) & 88.18 & 59.45 & 50.87 & 56.875 & 95.086 & 34.56 & 34.46 & 34.78 & 34.60 \\
    \mybottomrule
\end{tabular}
}
\end{table*}

\begin{table*}[t]
\centering
\captionsetup{font=small}
\caption{Extended experiments under severe spatial occlusion.}\label{tab:comp_mask_extend}
\setlength\tabcolsep{2.5pt}
\renewcommand{\arraystretch}{1}
\resizebox{1\textwidth}{!}{
\begin{tabular}{l|ccc|ccc|ccc|ccc|ccc}
\mytoprule
 \multirow{2}{*}{Methods / (\%)\textuparrow} & \multicolumn{3}{c|}{50 \texttimes~50} & \multicolumn{3}{c|}{100 \texttimes~100} & \multicolumn{3}{c|}{150 \texttimes~150} & \multicolumn{3}{c|}{200 \texttimes~200} & \multicolumn{3}{c}{250 \texttimes~250} \\
 \mycmidrule{2-16}
 & gACC & mACC & mIoU & gACC & mACC & mIoU & gACC & mACC & mIoU & gACC & mACC & mIoU & gACC & mACC & mIoU \\ 
 \mymidrule
 TokenFusion \cite{wang2022multimodal} & 88.03 & 62.54 & 52.50 & 84.92 & 59.90 & 48.00 & 79.63 & 56.34 & 43.85 & 75.25 & 52.08 & 39.76 & 73.27 & 49.68 & 37.39 \\
 CMX \cite{zhang2023cmx} & \cellcolor{third}90.53 & \cellcolor{third}70.73 & 57.76 & 86.43 & \cellcolor{third}67.40 & \cellcolor{third}53.73 & \cellcolor{third}82.24 & \cellcolor{third}64.12 & \cellcolor{second}49.59 & \cellcolor{third}80.21 & 59.21 & \cellcolor{second}45.96 & \cellcolor{second}79.33 & \cellcolor{third}56.81 & \cellcolor{second}43.79 \\
 CMNeXt \cite{zhang2023delivering} & \cellcolor{second}91.22 & \cellcolor{second}71.91 & \cellcolor{second}60.65 & \cellcolor{third}87.53 & \cellcolor{second}68.71 & 53.70 & 81.62 & \cellcolor{second}64.30 & 48.01 & 77.08 & \cellcolor{third}59.51 & 43.71 & 74.37 & 56.42 & 41.12 \\
 EISNet \cite{xie2024eisnet} & 90.43 & 68.12 & \cellcolor{third}57.89 & \cellcolor{second}88.20 & 66.19 & \cellcolor{second}54.47 & \cellcolor{second}83.83 & 63.31 & \cellcolor{third}49.30 & \cellcolor{second}80.99 & \cellcolor{second}60.71 & \cellcolor{third}45.92 & \cellcolor{third}78.83 & \cellcolor{second}58.78 & \cellcolor{third}43.51 \\
 \textbf{Ours} & \cellcolor{first}\textbf{92.46} & \cellcolor{first}\textbf{74.08} & \cellcolor{first}\textbf{64.53} & \cellcolor{first}\textbf{92.20} & \cellcolor{first}\textbf{72.91} & \cellcolor{first}\textbf{63.87} & \cellcolor{first}\textbf{91.05} & \cellcolor{first}\textbf{70.40} & \cellcolor{first}\textbf{61.32} & \cellcolor{first}\textbf{89.01} & \cellcolor{first}\textbf{66.90} & \cellcolor{first}\textbf{58.02} & \cellcolor{first}\textbf{87.26} & \cellcolor{first}\textbf{64.06} & \cellcolor{first}\textbf{54.95} \\
 \mybottomrule
\end{tabular}
}
\end{table*}

\begin{figure*}[t]
    \centering
    \includegraphics[width=1.0\linewidth]{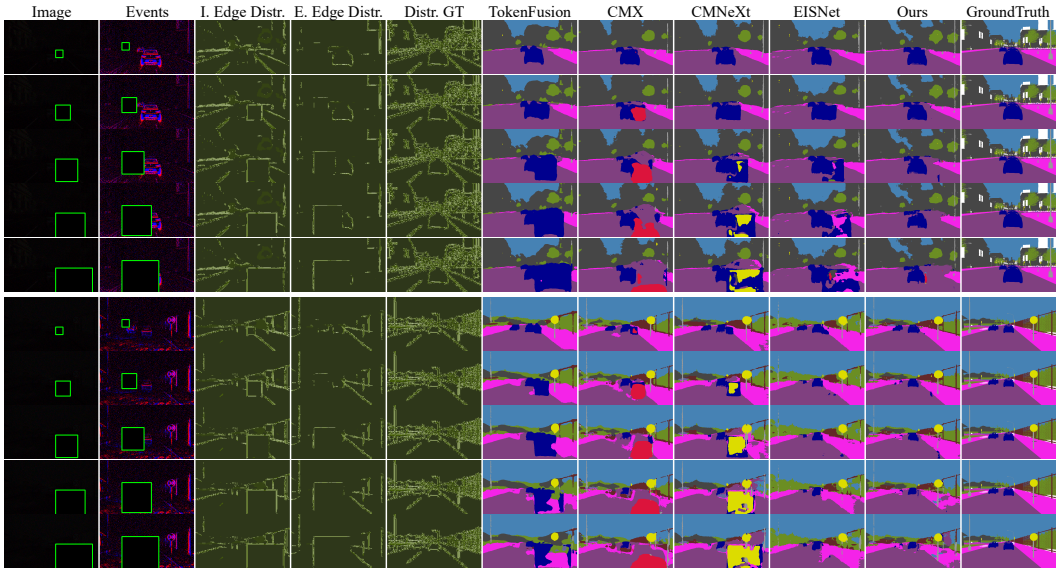}
    \captionsetup{font=small}
    \caption{Qualitative results of extended experiments under severe spatial occlusion on DERS-XS.}
    \label{fig:SUPP-MASK-Extend}
\vspace{1em}
\end{figure*}

\section{Extended experiments under severe spatial occlusion}
\label{sec:appendixE}

We extend our experiments under severe spatial occlusion. In the main text, we apply masking on RGB and event with mask areas size of 100 \texttimes 100. We further conduct experiments with masking areas of different sizes. Masking areas of size 50 \texttimes~50, 150 \texttimes~150, 200 \texttimes~200, and 250 \texttimes~250 are applied at coordinate \textlangle350, 200\textrangle~for RGB and \textlangle150, 150\textrangle~for event respectively. Both event and RGB are applied masking. The excess part is ignored if the masking area exceeds the spatial area.

\Cref{tab:comp_mask_extend} demonstrates the comparison results of extended experiments under severe spatial occlusion. As the size of masking area increases, the performance of all methods degrades, and our method consistently outperforms other multi-modality methods on different masking settings. CMX has a lower performance than CMNeXt when the masking areas are small, but it outperforms CMNeXt when the masking areas become larger. \Cref{fig:SUPP-MASK-Extend} demonstrates the qualitative comparison results of extended experiments on masking settings on DERS-XS. Under different masking settings, although all methods are affected by the modality imbalance and information loss caused by masking, our method obtains more reliable information based on uncertainty edge-aware joint optimization and edge consolidation, thus avoiding the misleading information of masking areas as much as possible.

\section{More qualitative comparison results}
\label{sec:appendixF}

\begin{figure*}[t]
    \centering
    \includegraphics[width=1.0\linewidth]{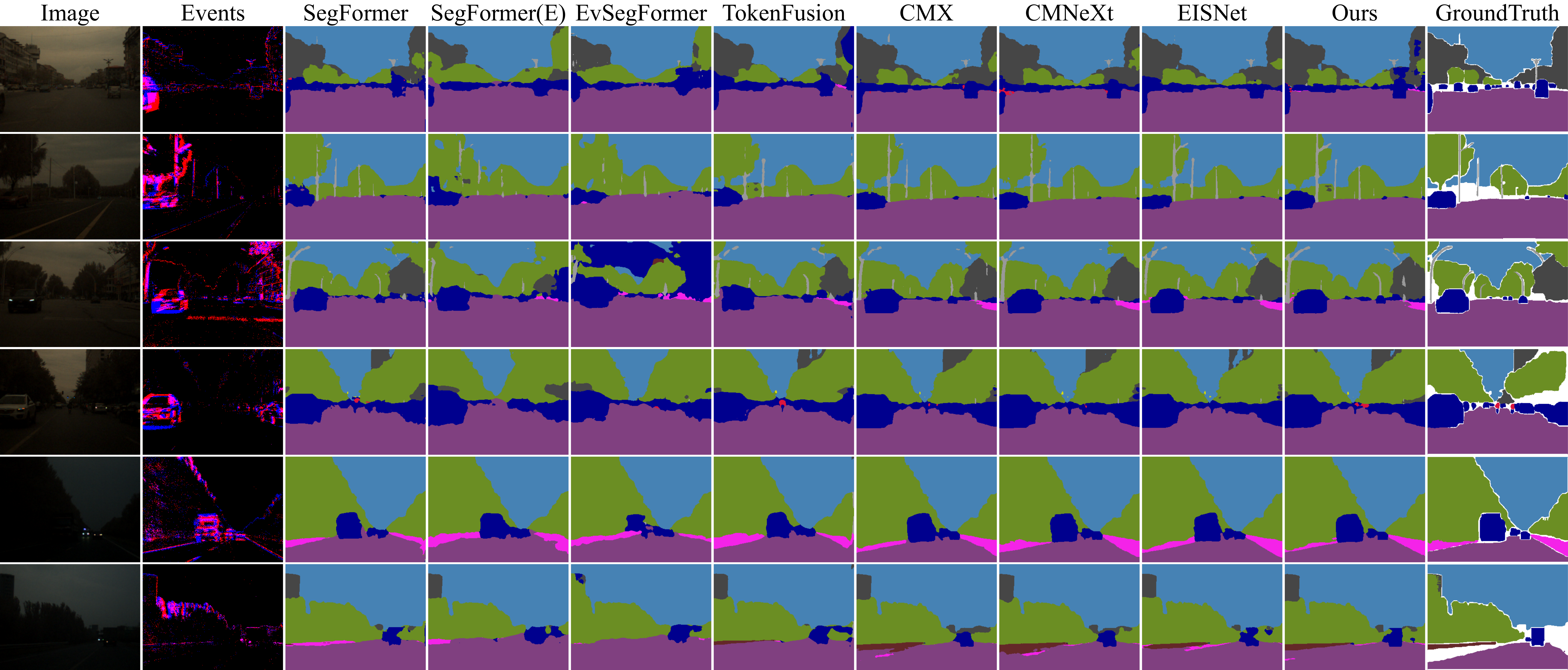}
    % \vspace{-2em}
    \captionsetup{font=small}
    \caption{More qualitative comparison results on DERS-XR.}
    \label{fig:SUPP-DERS-XR}
    % \vspace{-1.0em}
\end{figure*}

\begin{figure*}[t]
    \centering
    \includegraphics[width=1.0\linewidth]{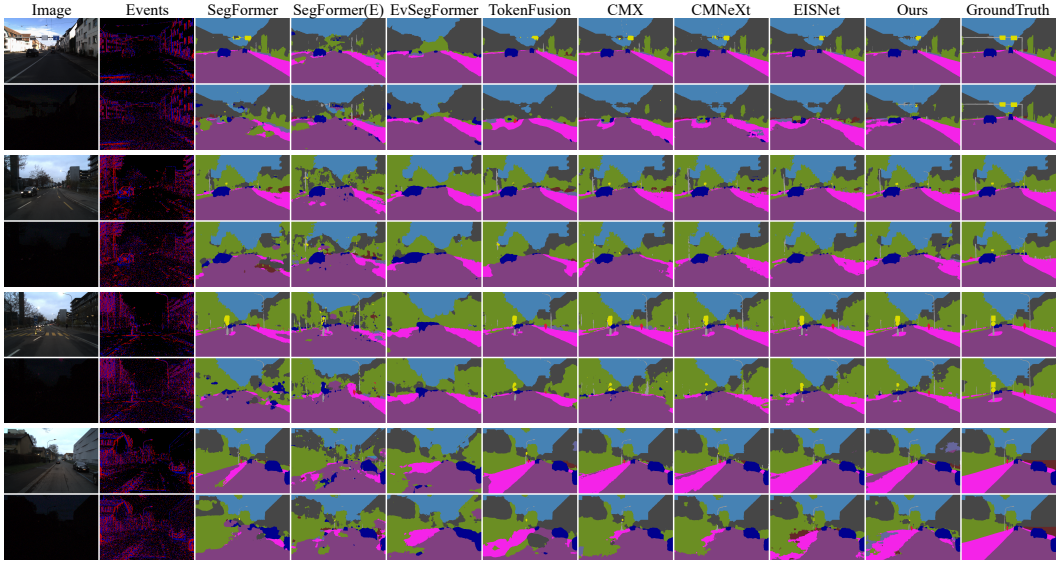}
    % \vspace{-2em}
    \captionsetup{font=small}
    \caption{More qualitative comparison results on DSEC-Semantic and DSEC-Xtrm.}
    \label{fig:SUPP-DSEC}
    % \vspace{-0.75em}
\end{figure*}

\begin{figure*}[t]
    \centering
    \includegraphics[width=1.0\linewidth]{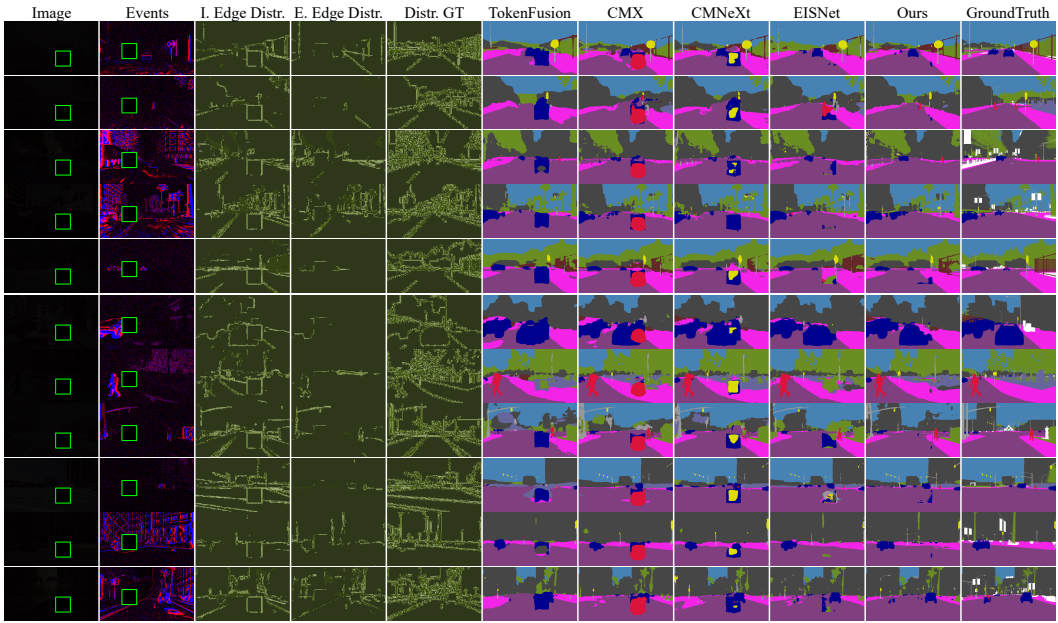}
    % \vspace{-2em}
    \captionsetup{font=small}
    \caption{More qualitative comparison results under E-RGB mask setting on DERS-XS.}
    \label{fig:SUPP-MASK}
    % \vspace{-1.0em}
\end{figure*}

This section is an extension of the qualitative results of the experiments in the main text. Figures are placed at the end of the appendix. \Cref{fig:SUPP-DERS-XS} demonstrates more qualitative comparison results on DERS-XS. \Cref{fig:SUPP-DERS-XR} demonstrates more qualitative comparison results on DERS-XR. \Cref{fig:SUPP-DSEC} demonstrates more qualitative comparison results on DSEC-Semantic and DSEC-Xtrm. \Cref{fig:SUPP-MASK} demonstrates more qualitative comparison results under E-RGB mask setting on DERS-XS. 

As shown in \cref{fig:SUPP-DERS-XS}, compared with other methods, our method can segment vehicles and pedestrians with complex contours more stably and robustly under extreme conditions. Especially for pedestrians, the contours of pedestrians segmented by other methods are not sharp enough, and sometimes even fail to detect the existence of pedestrians. our method can effectively locate pedestrians and segment the contours of pedestrians accurately.

As shown in \cref{fig:SUPP-DERS-XR}, our method performs well on real-world extreme scenes when fine-tuned with a small amount of real-world data. This also confirms that our simulated dataset DERS-XS can effectively provide prior knowledge when the amount of real-world data is small.

As shown in \cref{fig:SUPP-DSEC}, our method can handle Event-RGB semantic segmentation under normal conditions well, and when the modality information is lost under extreme conditions, our method is still able to identify the contours of ambiguous vehicles and pedestrians and maintains the segmentation performance better than other methods.

As shown in \cref{fig:SUPP-MASK}, our method can still achieve relatively accurate segmentation results even when the inputs are partially masked. The image edge distribution and events edge distribution indicate their different awareness of edges. The two modalities complement each other and are dynamically fused based on their confidences and uncertainties. The results show that our method is more robust and resilient than other methods in the cases of modality imbalance and failure.

\begin{figure*}[t]
    \centering
    \includegraphics[width=1.0\linewidth]{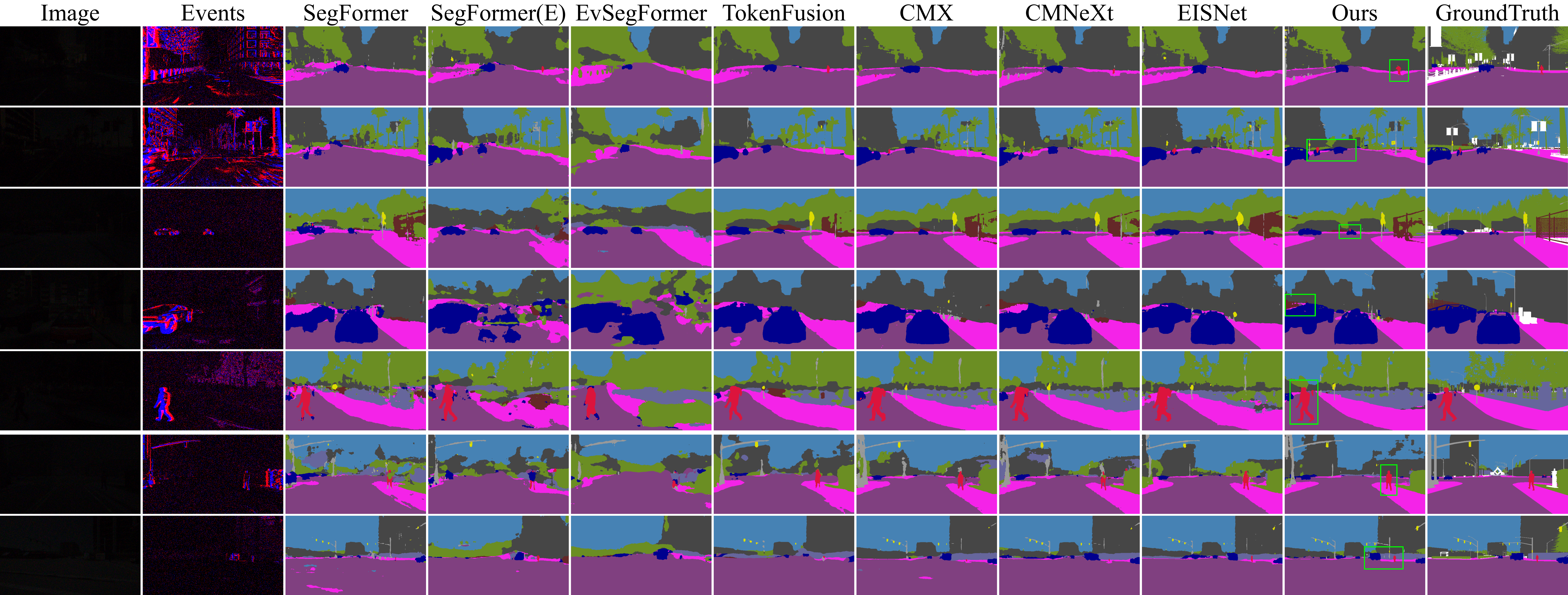}
    % \vspace{-2em}
    \captionsetup{font=small}
    \caption{More qualitative comparison results on DERS-XS.}
    \label{fig:SUPP-DERS-XS}
    % \vspace{-0.75em}
\end{figure*}

\end{document}